\documentclass[mnsc,nonblindrev]{informs3} 

\OneAndAHalfSpacedXI

\usepackage{natbib}
 \bibpunct[, ]{(}{)}{,}{a}{}{,}%
 \def\bibfont{\small}%
\usepackage{booktabs}
\usepackage{tikz}
\usetikzlibrary{positioning, arrows.meta, shapes.geometric}
\usepackage{multirow}
\usepackage{color}
\usepackage{graphicx}
\usepackage{float}
\usepackage{subfig}
\usepackage{amsmath}
\usepackage{tikz}
\usetikzlibrary{positioning, arrows.meta, shapes}
\usepackage{colortbl}  
\usepackage{xcolor}
\usepackage{indentfirst}
\usepackage{array} 
\usepackage{tikz}

\theoremstyle{definition}

\TheoremsNumberedThrough     
\ECRepeatTheorems

\EquationsNumberedThrough    

\begin{document}



\RUNTITLE{ Spatiotemporal  Traffic Forecasting Through Multimodal Network Integration}

\TITLE{ FusionTransNet for Smart Urban Mobility: Spatiotemporal  Traffic Forecasting Through \\ Multimodal Network Integration}


\ARTICLEAUTHORS{%
\AUTHOR{Binwu Wang\footnote{The two authors have equal contributions.}}
\AFF{University of Science and Technology of China} 
\AUTHOR{Yan Leng\footnotemark[1]}
\AFF{University of Texas at Austin}
\AUTHOR{Guang Wang}
\AFF{Florida State University}
\AUTHOR{Yang Wang}
\AFF{University of Science and Technology of China}
} 

\ABSTRACT{This study develops \textsf{FusionTransNet}, a framework designed for Origin-Destination (OD) flow predictions within smart and multimodal urban transportation systems. 
Urban transportation complexity arises from the spatiotemporal interactions among various traffic modes. 
Motivated by analyzing multimodal data from Shenzhen, a framework that can dissect complicated spatiotemporal interactions between these modes, from the microscopic local level to the macroscopic city-wide perspective, is essential. 
The framework contains three core components: the Intra-modal Learning Module, the Inter-modal Learning Module, and the Prediction Decoder. The Intra-modal Learning Module is designed to analyze spatial dependencies within individual transportation modes, facilitating a granular understanding of single-mode spatiotemporal dynamics. The Inter-modal Learning Module extends this analysis, integrating data across different modes to uncover cross-modal interdependencies, by breaking down the interactions at both local and global scales. 
Finally, the Prediction Decoder synthesizes insights from the preceding modules to generate accurate OD flow predictions, translating complex multimodal interactions into forecasts.
Empirical evaluations conducted in metropolitan contexts, including Shenzhen and New York, demonstrate \textsf{FusionTransNet}'s superior predictive accuracy compared to existing state-of-the-art methods. 
The implication of this study extends beyond urban transportation, as the method for transferring information across different spatiotemporal graphs at both local and global scales can be instrumental in other spatial systems, such as supply chain logistics and epidemics spreading. 
}


\maketitle


%


	\section{Introduction}
    \label{sec:introduction}
The accelerating pace of urbanization poses significant challenges and opportunities for urban management and policy-making, particularly in the domain of transportation~\citep{zhang2023ridesharing}. As cities grow and evolve, managing urban mobility becomes increasingly complex, requiring sophisticated analytical tools to decode patterns of movement and predict future traffic flows~\citep{wang2022deep, ashok2002estimation,bachir2019inferring, yang2021novel, ketter2023information}. In this context, effective policy-making hinges on the integration of diverse data sources to craft resilient and sustainable urban environments. This challenge is at the heart of advancing smart urban systems, aligning with the United Nations' Sustainable Development Goals, especially those focused on creating inclusive, safe, resilient, and sustainable cities.\footnote{\url{https://sdgs.un.org/goals}} 
    Our study, focusing on spatiotemporal learning for OD  flow prediction, aims to contribute to these broader objectives by offering travel demand insights to  support efficient planning and operation of urban transportation systems.

The complexity inherent in urban transportation, particularly with the advent of multimodal travel journeys (e.g., transferring from bike sharing to subways), delineates both a significant challenge and a promising opportunity for the refinement of OD flow predictions. 
Current research showcases considerable advancements in understanding single-mode transportation flows but often falls short of fully addressing the dynamics and inter-modal interactions characteristic of modern urban mobility systems~\citep{wang2022deep, zhao2019predicting}. Moreover, the limited exploration of multimodal data within existing literature tends to adopt a coarse-grained approach, bypassing the nuanced spatial and temporal dependencies that are paramount at both local and global scales~\citep{ye2019co, xu2022adaptive}. This observation underscores a critical research gap: the need for a comprehensive framework that effectively integrates multimodal transportation data to enhance OD flow predictions through capturing the complex inter-modal interactions and spatiotemporal dependencies inherent in urban multimodal transportation systems.

\begin{figure}[!htbp]
		\centering
	
       \subfloat[\label{fig:commercial1}]{\includegraphics[width=0.35\linewidth]{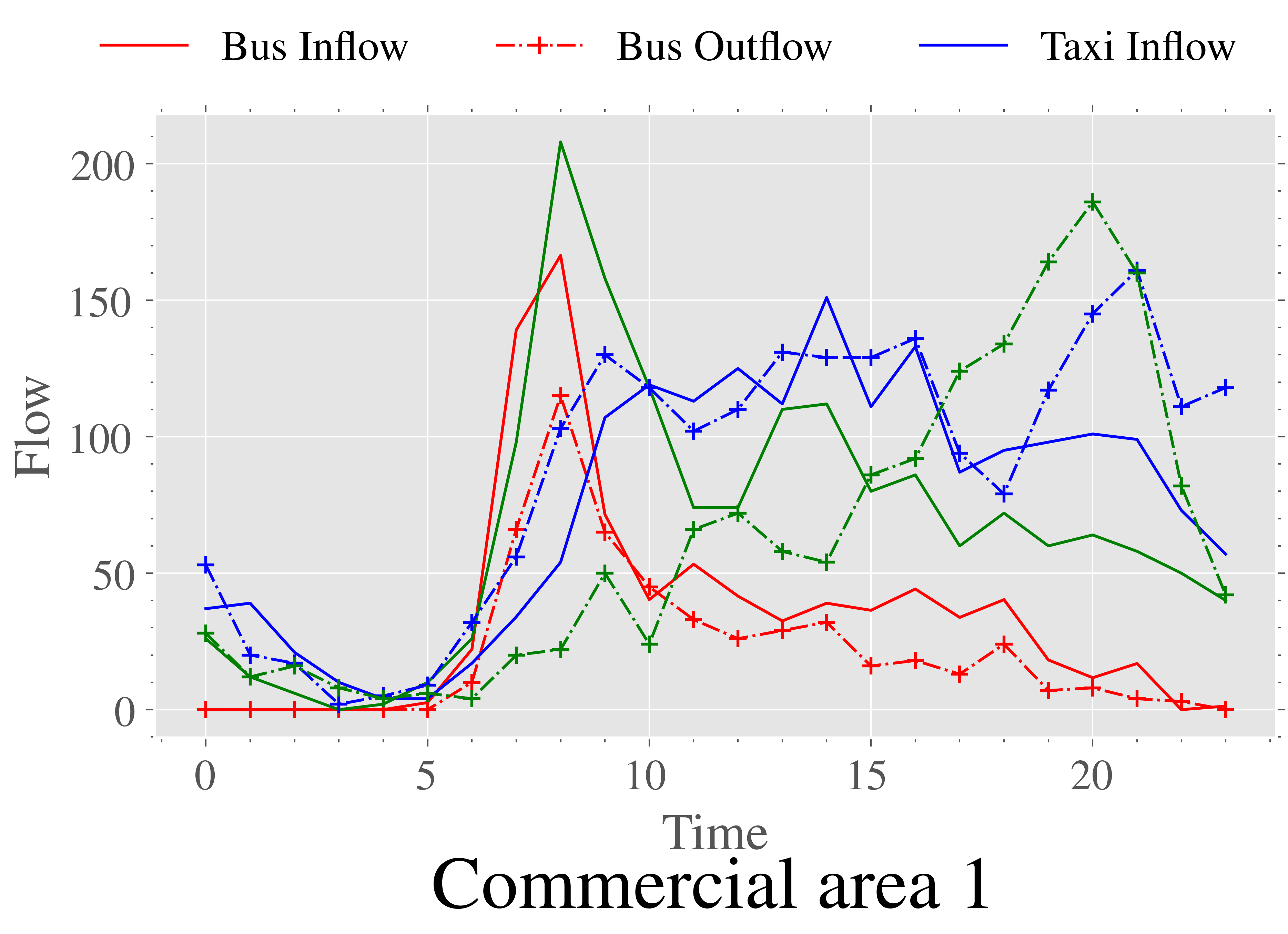}}
       \subfloat[\label{fig:futian}]{\includegraphics[width=0.35\linewidth]{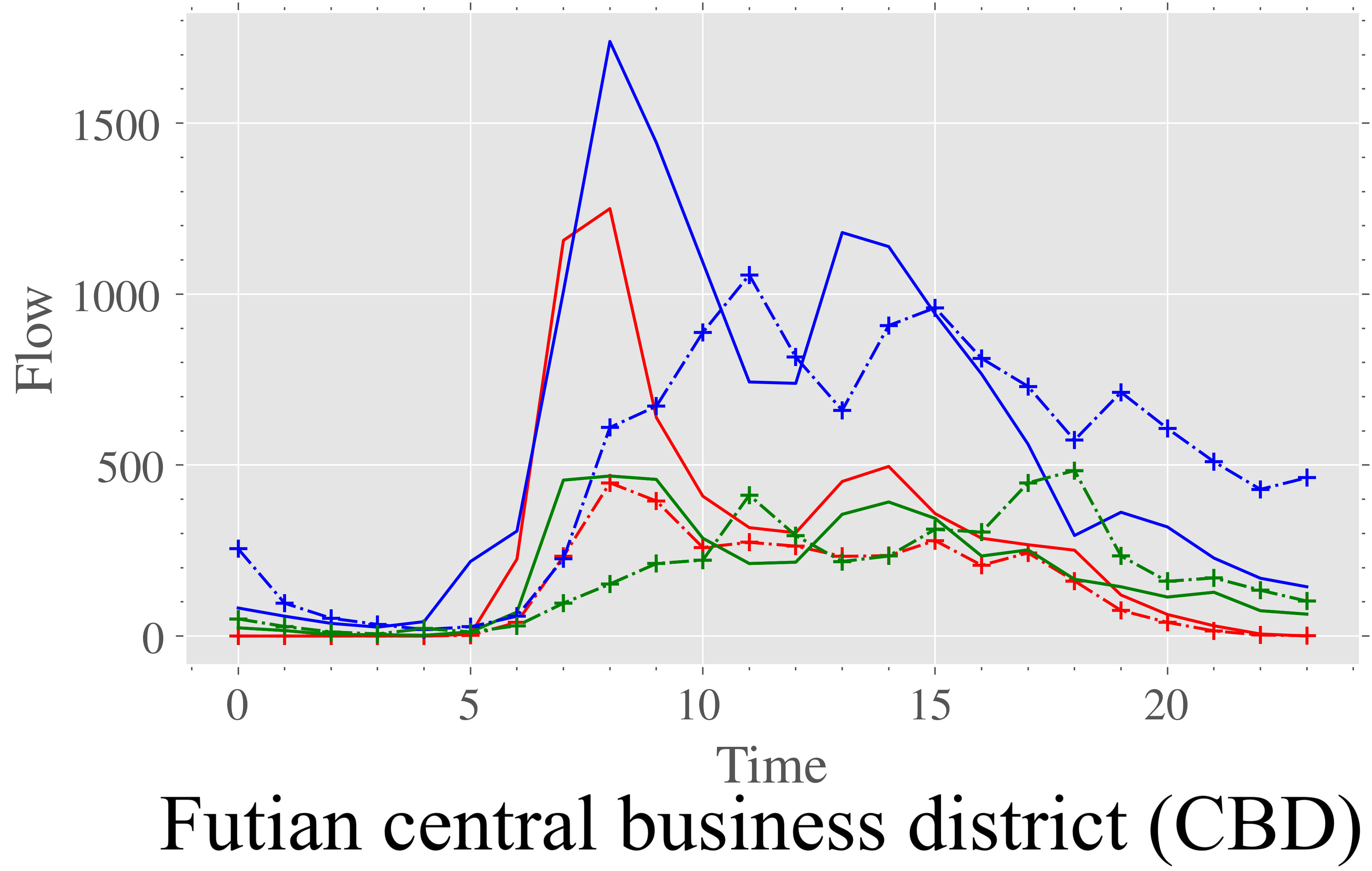}}
     \subfloat[ \label{fig:resi1}]
     {\includegraphics[width=0.35\linewidth]{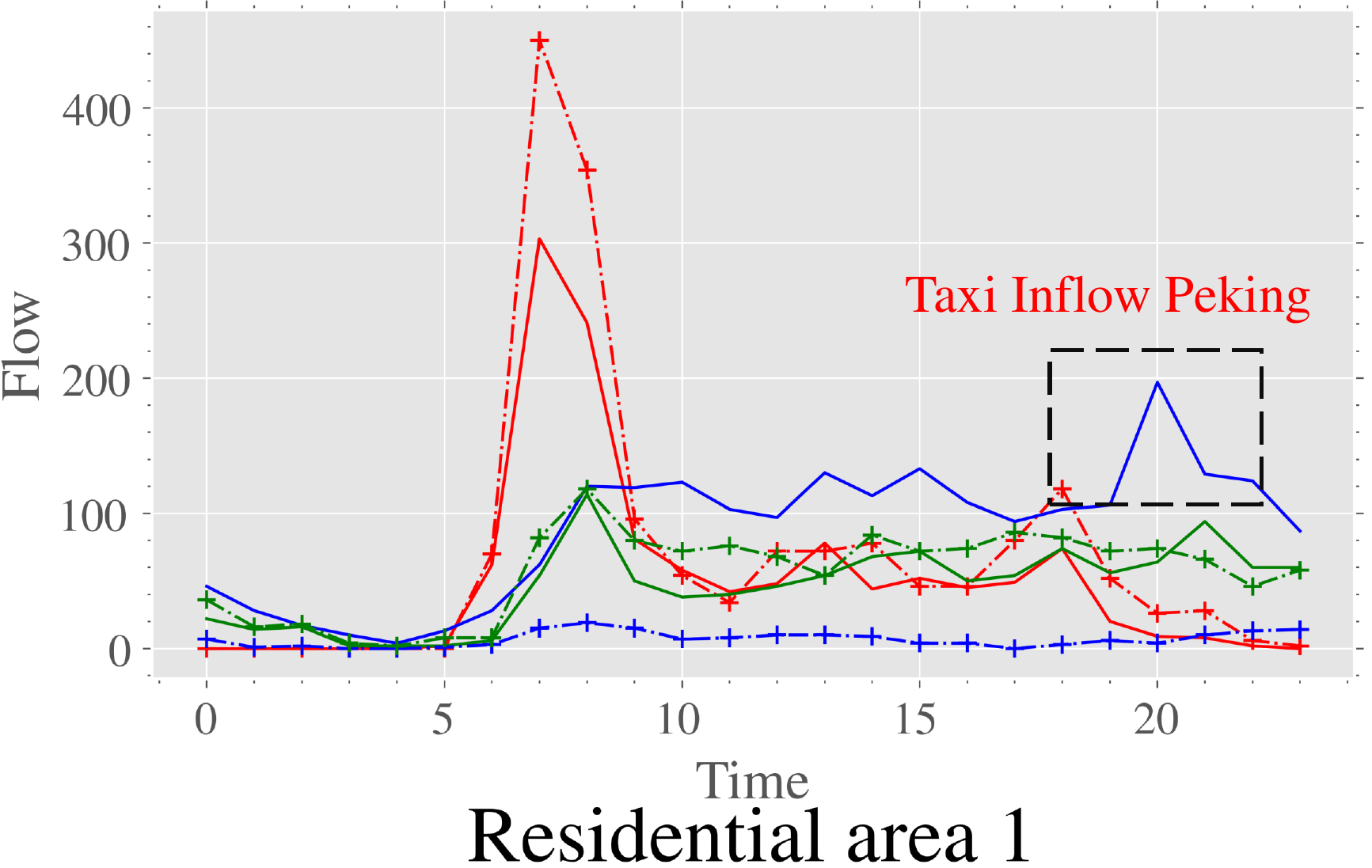}}\\
       \subfloat[\label{fig:commercial2}]{\includegraphics[width=0.35\linewidth]{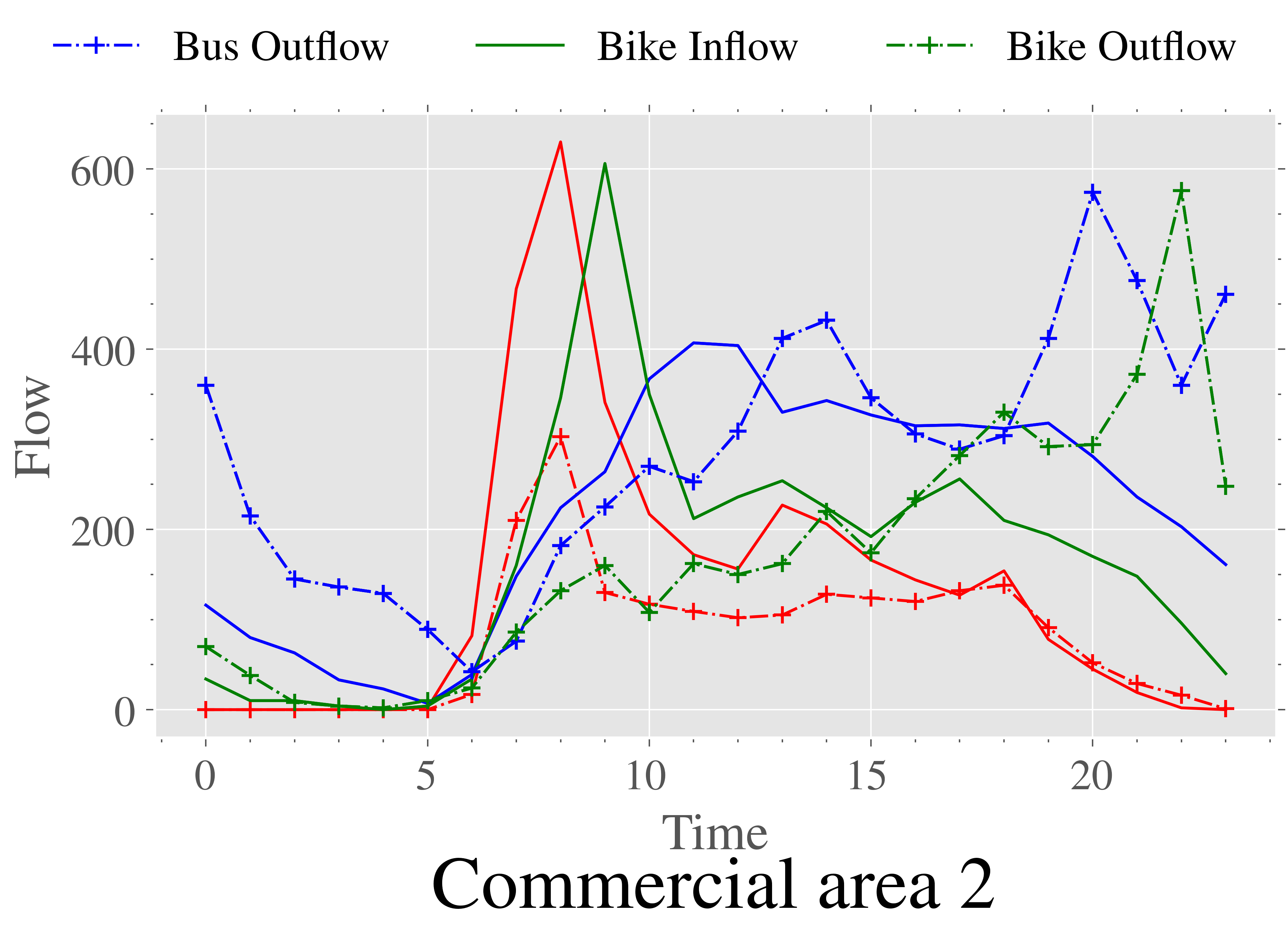}}
           	\subfloat[ \label{fig:north}]
            {\includegraphics[width=0.35\linewidth]{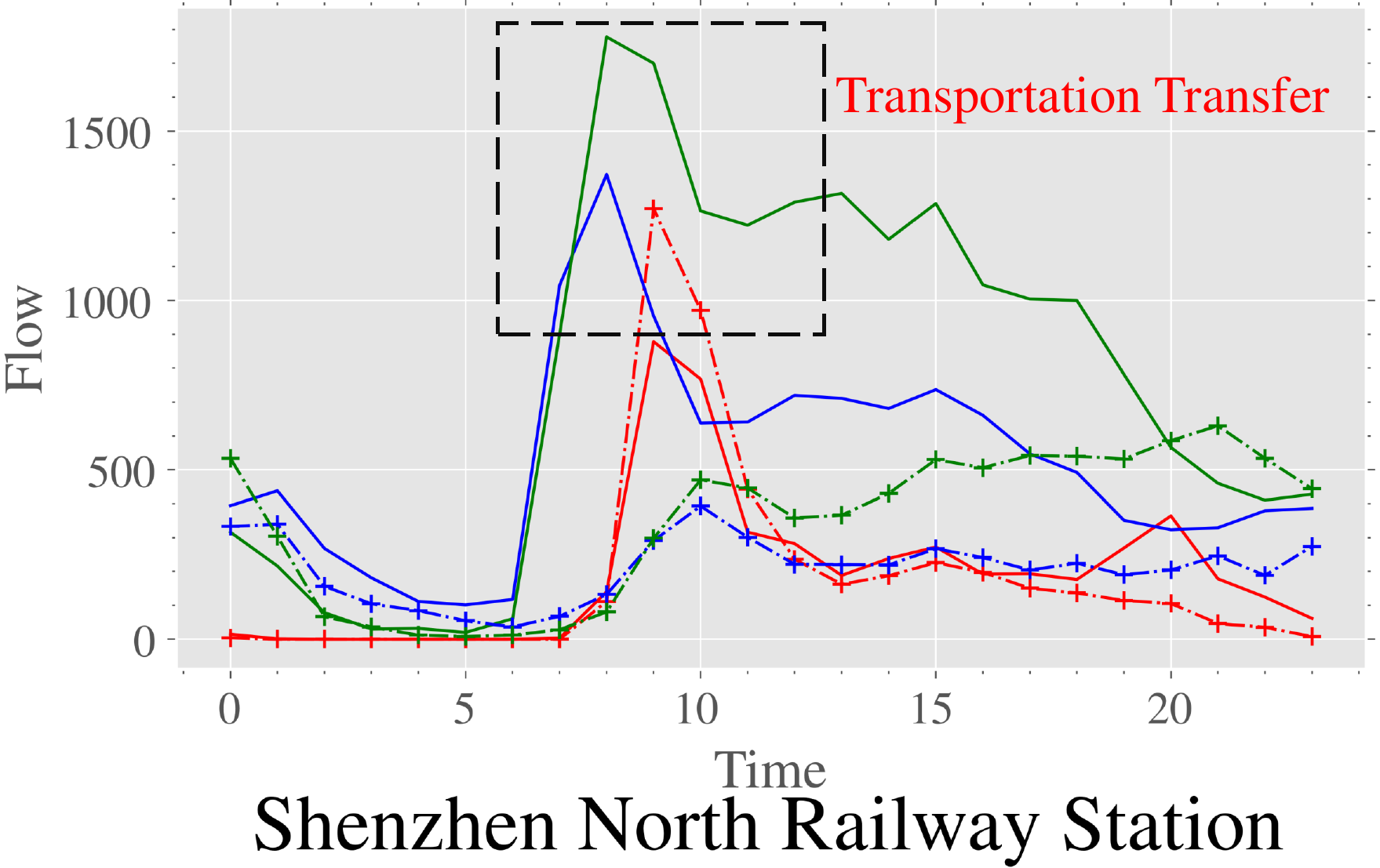}}
            \subfloat[\label{fig:resi2}]
   {\includegraphics[width=0.35\linewidth]{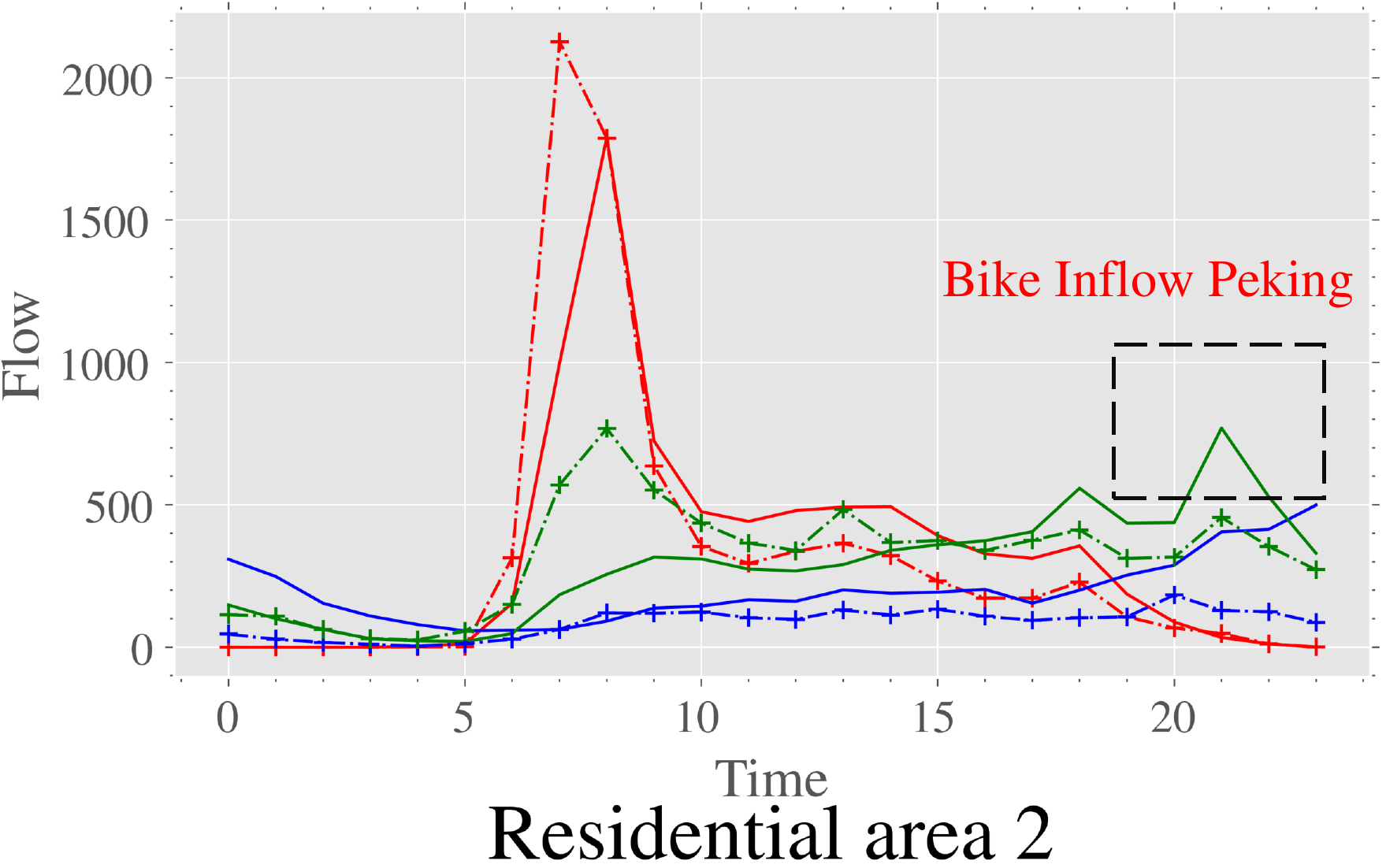}}
		\caption{Motivating examples: Hourly traffic distribution of six regions in Shenzhen.} 
         \label{fig:pattern}
	\end{figure}

Our motivation to integrate multiple transportation modes for OD flow predictions originates from a preliminary multimodal transportation data analysis.\footnote{
In our study, the term `modal' or `mode' specifically denotes transportation modes such as bus, taxi, or bikes, in accordance with the convention in transportation literature. This usage differs from the concept of multimodal data found in certain streams of machine learning literature, where it typically refers to the combination of text and image data. 
}
We use data from the technology hub in China, Shenzhen\footnote{Shenzhen's rapid urbanization, technological innovation, and diverse transportation infrastructure make it an ideal case study to examine the complexities of urban mobility and the implementation of smart urban systems. Additionally, its global significance as a hub for technology and manufacturing offers valuable insights into leveraging information systems and data-driven technologies to address urban mobility challenges effectively.}, to illustrate the predictive values and the complicated spatial patterns inherent in cross-modal and intra-modal traffic patterns (in Figure~\ref{fig:pattern}). For example, in Figures~\ref{fig:commercial1}-\ref{fig:commercial2}, peak flows in commercial zones during morning hours suggest a transferable predictive power between bus and taxi inflows. However, the asymmetry in taxi outflows in Figure~\ref{fig:commercial1} cautions against a simplistic transference of inflow patterns to outflows. In Figure~\ref{fig:futian}, the Futian CBD's early peak in bus inflows contrasts with the pattern observed at Shenzhen North Railway Station in Figure~\ref{fig:north}, where a later spike in bike inflows reflects distinct last-mile travel behaviors. Figures~\ref{fig:resi1}--\ref{fig:resi2} in residential areas further emphasize the diversity of regional patterns, with Figure~\ref{fig:resi1}'s evening taxi peak and Figure~\ref{fig:resi2}'s bike preference hinting at modal interdependencies that differ from commercial areas. 
These findings underscore the richness of spatiotemporal interactions and the necessity of accounting for regional specificities and inter-modal relationships in predictive modeling. 

Further compounding the need for our study, existing methodologies in urban transportation modeling exhibit significant limitations in capturing the full spectrum of multimodal interactions and the nuanced spatiotemporal dependencies crucial for accurate OD flow prediction. Predominantly, these models provide an isolated view of traffic flows, aggregating data into broad inflow or outflow metrics without a detailed examination of how different modes of transportation—taxis, buses, bikes—interact and influence each other across diverse urban landscapes. This simplified approach overlooks the essential variability in traffic patterns driven by time of day, specific urban zones, and the unique behavior of commuters navigating these spaces.

In addressing the complexities of urban mobility and the predictive challenges associated with multimodal transportation systems, our study introduces the \textsf{FusionTransNet}, a novel multimodal spatiotemporal learning framework designed to enhance OD flow predictions. This framework is structured around three components: an Intra-Modal Learning Module an Inter-Modal Learning Module, and a final Prediction Decoder. 

The framework begins with the Intra-Modal Learning Module, where spatiotemporal learning blocks, including two spatiotemporal graph convolutional network modules for capturing spatial dependencies within each individual transportation mode. This phase is critical for understanding how traffic conditions in one area can influence adjacent areas, serving as the building block for accurate spatial analysis. This is complemented by temporal attention aimed at identifying important temporal patterns, essential for forecasting traffic conditions over time. 
The most unique component in our framework is the Inter-Modal Learning Module. This module consists of global and local fusion strategies to model dependencies across different spatial resolutions and modalities accurately. The global fusion strategy focuses on capturing broad spatial relationships important for understanding city-wide traffic patterns, while the local fusion strategy aims to capture detailed spatial dependencies relevant at a micro level, such as specific streets or neighborhoods. Additionally, a multiple perspective interaction sub-module integrates features from various modalities, improving the representation of each node (or spatial unit) by incorporating information from multiple sources. This module's role is crucial in synthesizing data across modalities, offering a comprehensive view of urban traffic patterns. 
Our framework concludes with a Prediction Module that utilizes the learned embeddings from the Intra- and Inter-modal Learning Modules to forecast future OD flows.

	Our paper makes the following contributions. First, we develop 
    \textsf{FusionTransNet}, a  framework that combines multimodal dynamic traffic networks for enhancing OD flow predictions. 
    Our  dual fusion strategies—global and local—in the inter-modal learning module ensure comprehensive coverage of spatial dependencies across modes at different spatial locations and temporal scales. These components are instrumental in modeling the complex spatiotemporal dependencies critical for accurate multimodal traffic forecasting.
    
    Second,  we conceptualize urban mobility as a series of interconnected spatiotemporal networks, each representing a different transportation mode, within a unified  framework. This approach expands upon the single-mode transportation analysis \citep{wang2022deep, zhao2019predicting, zhou2021urban, zhao2022hyper} by providing a granular model of how various transportation modes are interconnected at both a macroscopic city-wide level and a microscopic, local scale, across varying temporal resolutions.
    Our model facilitates an in-depth examination of the interactions between various modes of transportation, offering a detailed and dynamic view of urban mobility. 
    Beyond urban transportation, the conceptual framework of interconnected (spatiotemporal) graphs is highly relevant for modeling complex networks in other domains, such as supply chain networks. 
   
   Third, through empirical evaluation in two metropolitan contexts (Shenzhen and New York), we validate the effectiveness of \textsf{FusionTransNet}, and demonstrate its superiority over existing benchmarks. The significant improvements in predictive accuracy—4.39\% to 7.17\% on the Shenzhen dataset—underscore the practical impact and scientific advancement our model represents in the domain of OD flow prediction methodologies.
    We also performed ablation studies and interpretation analyses to demonstrate the significance of the each component of our framework. 
    These components can be used to intricately model other complex systems with interconnected networks.

        \section{Related Work}
        
        The transition towards smart and sustainable urban mobility underscores the role of IS research in addressing the challenges within the transportation systems. \citet{kahlen2023smart}, \citet{lee2019impact}, and \citet{ketter2023information} highlight the importance of developing IS methodologies to improve transportation systems, emphasizing the shift towards intelligent and environmentally friendly mobility solutions. These studies advocate for a multidisciplinary and data-driven approach that leverages multimodal transportation data, positioning our research within the context of enhancing urban mobility solutions. This perspective aligns with the growing recognition of the need for innovative strategies to address the complexities of urban transportation, suggesting a direct relevance of IS research to method developments to support  smart, sustainable transit systems.

        In this section, we review literature on traffic flow predictions, which can be generally classified into three categories: Shallow learning methods, traditional Deep Learning-based methods, and Graph Neural Network (GNN)-based methods, which we review in this section. 

        \subsection{Shallow Learning for OD Predictions}
        OD flow prediction plays an instrumental role in intelligent transportation systems~\citep{li2007generalized}. 
        The OD matrix provides a picture of the spatial distribution of the traffic. 
        It is the second stage of the traditional four-step travel model (i.e., trip generation, trip distribution, mode choice, and route assignment) for determining transportation forecasts~\citep{mcnally2007four}. 
        The four-step model is a widely-adopted model in the 20th century when data is limited. 
        Traditionally, trip distribution matches trip generation from the first stage -- based on the frequencies of origins and destinations of trips in each zone, often using the seminal gravity model. 
        The gravity model takes into account the relative activity at the origin and destination center, as well as the travel cost between them. 

        As more data became available, researchers started to use statistical methods. Approaches like time-series models, autoregressive integrated moving average (ARIMA), and exponential smoothing have been utilized to predict OD flows~\citep{xue2015short}. These models were primarily favored for their interpretability and straightforward application. 
        In the meantime, researchers employed linear regression methods to understand relationships between variables and predict future traffic flows~\citep{williams2003modeling}. These models presented the advantage of easy implementation and direct interpretation of the effects of predictors on OD flows.
        
        As machine learning becomes available, more recent methods have been used. Kernel-based methods, particularly the Support Vector Machine (SVM), have been applied for OD time-series prediction~\citep{sapankevych2009time}. SVMs map input data into a high-dimensional space and find an optimal hyperplane that can be used for regression or classification. By leveraging the kernel trick, SVMs showcased their efficacy in capturing nonlinear patterns in traffic data without the explicit need for transformation.
        In addition, the Random Forest algorithm was often favored due to its robustness against noise and potential for feature importance evaluation. Research demonstrated that Random Forests often outperformed individual decision trees and linear models in OD flow prediction tasks~\citep{leshem2007traffic}.

        \subsection{Traditional Deep Learning for OD Prediction}

        OD flow prediction has witnessed significant advancements through the adoption of deep learning techniques, reflecting the broader trend in smart cities and intelligent transportation systems. 
        The overarching aim has been to devise algorithms and models that can intelligently predict movement patterns, thereby optimizing traffic management and reducing congestion~\citep{ashok2002estimation,bachir2019inferring,li2017hybrid,bera2011estimation,hazelton2000estimation}.
            
        A slew of methodologies have emerged over the past years, leveraging the capabilities of Convolutional Neural Networks (CNNs) and Long Short-Term Memory (LSTM) networks. MultiConvLSTM ~\citep{chu2019deep}  considered localized travel demands as image pixels and uses CNN with LSTM to learn spatiotemporal correlations.   They introduced an innovative data structure named the OD tensor to represent OD flows. This innovation was crucial as it catered to the high-dimensional attributes of OD tensor. In another work, the Contextualized spatiotemporal Network (CSTN)~\citep{liu2019contextualized} was formulated, where CNNs were employed to discern both local and global spatial relationships, and LSTMs were integrated for grasping the time series evolution. 
        Despite the popularity of CNN in these methods, CNN cannot effectively handle graph-structured data. While these models offered substantial advancements in OD prediction, certain limitations persisted. CNNs, despite their efficacy in image and sequence data, struggled with graph-structured data. Recognizing this shortcoming, ~\citet{noursalehi2021dynamic} developed a novel approach with the Multi-Resolution Spatiotemporal Neural Network (MRSTN). The uniqueness of MRSTN lies in its use of discrete wavelet transform, ensuring a multi-resolution requirement decomposition, thereby efficiently capturing both spatial and temporal dependencies.

        \subsection{Graph Neural Networks for OD Prediction}
        Recent progresses in graph neural networks (GNN), with graph convolutional networks (GCNs) and graph attention networks (GAT) as prominent examples, opened up new possibilities for modeling graph data with pronounced spatial capabilities~\citep{liu2017novel, munizaga2012estimation, du2019deep}. 
        
        Researchers develop deep learning models building upon GCNs to the OD prediction task to achieve better forecasting performance. \citet{ZHANG2021102851} constructed a dynamic OD graph to describe the ride-hailing demand data, and proposes a neural structure called Dynamic Node Edge Attention Network (DNEAT) to solve the challenge of OD demand prediction in the ride-hailing service platforms MPGCN~\citep{shi2020predicting} and ODCRN~\citep{jiang2021countrywide} designed an extended two-dimensional form of GCN based on two-dimensional discrete Fourier transform to encode OD flow matrix, and LSTM module is used to learn temporal correlation. DNEAT~\citep{ZHANG2021102928} designed a spatiotemporal attention network and exploit different time granularity to mine complex temporal patterns. HMOD~\citep{zhang2022dynamic} is a dynamic graph representation learning framework for OD demand prediction, which integrates discrete time information and continuous time information of OD demand. CMOD~\citep{han2022continuous} is a continuous-time dynamic graph representation learning framework and proposes a hierarchical message passing module to model the spatial interactions of stations.
        Building upon GNN-based methods, \citet{wang2022deep} advanced traffic flow prediction by processing OD matrices by integrating CNNs and LSTMs, augmented by an attention mechanism and contextual embeddings for spatial and temporal analysis. Their method dynamically integrates LSTM outputs with contextual information to refine predictions. 
        
        Despite these important progresses, a notable gap remains. The majority of extant methodologies are primarily sculpted for single transportation modal OD flow predictions. A relatively uncharted area lies in the domain of multimodal transportation OD flow predictions. Tackling such predictions is challenging, primarily due to the complicated interaction dynamics across varied transportation modalities, both in spatial and temporal dimensions. This underscores a pressing need for novel methods capable of effectively addressing this multimodal challenge, in which there is a need to model both granular and macro dependencies. 
        We extend \citet{wang2022deep} by building interconnected multi-graphs to model urban systems more granularly. 
        We specifically incorporate both fusion strategies (global and local) for a deeper analysis of spatial dependencies and introducing the \textsf{ModeDistinctNet} to address mode-specific variabilities, enhancing prediction accuracy.  

        The application of deep learning extends beyond OD prediction in IS.  
        In addition to the above-mentioned studies, deep learning methods have been developed across diverse IS domains to solve important problems, from disaster management~\citep{zhang2023proactive} and privacy preservation~\citep{macha2023personalized} to marketing and consumer behavior prediction~\citep{sun2022predicting, chen2023attending}.  

 	\section{Model}

    \subsection{Problem Description}
    \label{subsec:problem}
    In this section, we discuss the key definitions and terms.
    We aim to predict the hourly travel demand between each origin-destination pair for each of the three transportation modals at time step $t \in \{1, ..., T\}$. 
    We partition an area of interest (e.g., a city) into a set of $P \times Q$ disjoint units, where $P - 1$ and $Q - 1$ are the number of horizontal and vertical splits. 
    This partition results in $P \times Q$ equally-sized \emph{spatial grids}.
    Each spatial grid is considered as $r_{pq}$, for $p \in \{1, ..., P\}$ and $q \in \{1, ..., Q\}$. 
    The origin and destination of each transportation modal take place in each spatial grid $r_{pq}$, including bike and taxi, which are station-free, and bus, which is station-based.  

    We then define $M$ sets of \emph{spatial units} ($\mathcal{V}_m$), one for each transportation modal $m$. 
    In our case, $m \in \mathcal{M} = \{\text{bus}, \text{taxi}, \text{bike}\}$ and $M = |\mathcal{M} |  3$. 
    Each $\mathcal{V}_m$ is a subset of spatial grid where passengers can depart from or arrive at this spatial grid using transportation modal $m$. 
    The cardinality of virtual station belonging to modal $m$ is  $|\mathcal{V}_m|$ where $|\mathcal{V}_m| = N_m$. Lastly, we define spatial units where there are overlapping services of multiple modals as \emph{multi-modal units}, and  $\mathcal{T}_n=[m_1,...,m_i]$ represents the traffic modal in the multi-modal unit $v_n$, which is a subset of $\mathcal{M}$.

    We next define Origin-Destination (OD) graph. 
     
    \begin{definition}[Origin-Destination graph]{An Origin-Destination (OD) graph for transportation modal $m$ at time step $t$ is denoted as $\mathcal{G}_m^t=(\mathcal{V}_m, \mathcal{E}_{m}^t, \mathbf{X}_{m}^t, \mathbf{M}_{m}^t)$, where $\mathcal V_m$ is the set of spatial units and the size of $|\mathcal{V}_m|$ is $N_m$.
    $\mathcal{E}_t^m$ is the set of edges. 
    If there exists non-zero traffic flow from spatial unit $i$ to $j$, then $(i, j) \in \mathcal{E}^t_m$. 
    We denote $\mathbf{M}^t_m \in \mathbb{R}^{N_m \times N_m}$ as the \emph{OD flow matrix}, where $\mathbf{M}^t_{m}\left[i,j\right]$ is the traffic flow from spatial unit $i$ to $j$ using transportation modal $m \in \mathcal{M}$. 
    $\mathbf{X}^t_m \in \mathbb{R}^{N_m \times k}$  represents the $k$-dimensional nodal feature, including information such as inflow, outflow, and types of points of interests (POIs).}
    \end{definition}


 
    \begin{problem}[Multimodal spatiotemporal learning for OD matrix prediction]
    {
    For a given focal transportation modal $m$, we predict the OD matrix $M_m^{t+1}$ at time step $t + 1$. 
    We take the OD graphs of the previous $L$ time steps historical data as input: $\{\mathcal{G}^t_i\}_{t-L}^{t-1}$ for $i \in \mathcal{M}$. 
    We output a mapping from this input to the OD matrix for transportation modal $f$ in time step $t + 1$, 
     \begin{equation}
    \mathbf{M}_m^{t+1}=\mathcal{F}\left(\{\mathcal{G}^t_i\}_{t-L}^{t-1}, \text{for }\; i \in \mathcal{M}\right), 
    \end{equation}
    where $\mathcal{F}(\cdot)$ is the mapping to be learned by our proposed model, and $\mathbf{M}_m^{t+1}$ corresponds to the OD matrix of the target transportation modal, where $m \in \mathcal{M}$, and $\mathcal{M} = \{\text{bus}, \text{taxi}, \text{bike}\}$. 
}
    \end{problem}

    Multimodal spatiotemporal learning for OD flow prediction is an extension of the traditional OD prediction task. 
    Conventional single-mode OD prediction can be formulated as $\mathbf{M}_m^{t+1}= \mathcal{F}\left(\{\mathcal{G}^t_m\}_{t-L}^{t-1} \right)$ for a single transportation mode $m$. 
    Multimodal spatiotemporal learning (and our focus) aims to exploit the predictive power of multiple transportation modes $\mathcal{M}$ to  predict the OD flow matrix of a focal transportation modal.

\begin{figure}
\centering
\begin{tikzpicture}[
    node distance=1cm and 0.5cm,
    mynode/.style={
        draw, align=center, text width=5.8cm, 
        rectangle split, rectangle split parts=3, rounded corners,
        rectangle split part fill={blue!20, orange!30, green!20} 
    },
    arrow/.style={-Stealth},
    mylabel/.style={align=center}
]

\centering 
\node[mynode] (intra) {
    \textbf{Intra-modal Learning}
    \nodepart{two}Goal: Capture temporal and spatial dependencies within modes
    \nodepart{three}OD-Adaptive GCN on ST-Graphs\\Temporal Attention Network
};

\node[mynode, right=of intra] (inter) {
    \textbf{Inter-modal Learning}
    \nodepart{two}Goal: Integrate cross-modal patterns at both Macroscopic and Microscopic Levels
    \nodepart{three}Global-Fusion Strategy\\Local-Fusion Strategy\\Multiple Perspective Interaction
};

\node[mynode, text width=3.9cm, right=of inter] (pred) { 
    \textbf{Prediction Decoder}
    \nodepart{two}Goal: Generate accurate OD flow predictions
    \nodepart{three}LSTM with learned multi-modal embeddings
};
\node[draw=none, rectangle split part fill={blue!20, orange!30}, text width=3.5cm, right=of inter] at (pred) {}; 

\draw[arrow] (intra) -- (inter);
\draw[arrow] (inter) -- (pred);

\end{tikzpicture}
\caption{The structure of \textsf{FusionTransNet}, illustrating its components and goals.}
\label{fig:skeleton}
\end{figure}
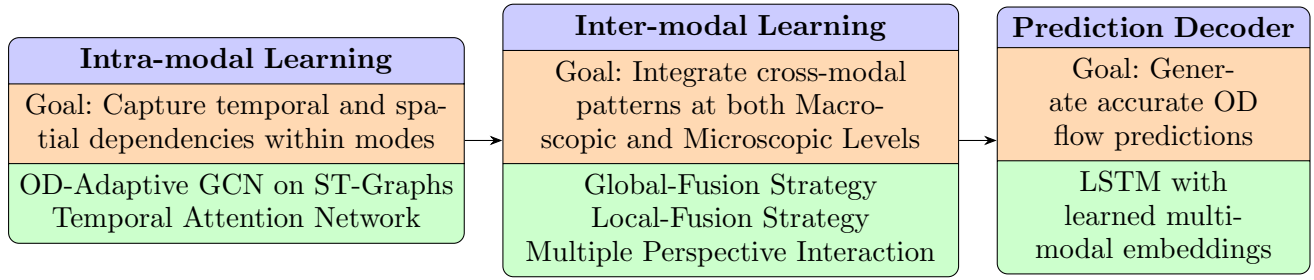

    \subsection{Overview of the Proposed Learning Framework: \textsf{FusionTransNet}}
    \label{subsec:framework}
	In this section, we describe our multimodal spatiotemporal learning framework---\textsf{FusionTransNet}. 
  \textsf{FusionTransNet} includes three  components: the Intra-Modal Learning Module, the Inter-Modal Learning Module, and a Prediction Decoder Module to predict the OD flow. 
    We provide a high-level skeleton of \textsf{FusionTransNet} in Figure \ref{fig:skeleton}. 
 
    \paragraph{Phase 1 (Intra-Modal Learning Module).} In the intra-modal learning phase, a spatiotemporal learning block that contains two convolutional networks on spatiotemporal graphs (OD-STGCN) is developed. 
    Spatially, OD-STGCN learns intra-modal spatial dependencies within each mode at the urban scale. 
    Temporally, the temporal attention network is utilized to learn temporal correlations within each transportation mode. 
        \begin{itemize}
            \item  Learning intra-modal spatial patterns and dependencies: Modeling spatial dependencies is important because spatial phenomena are often interrelated. For instance, the traffic condition on one neighborhood can significantly affect adjacent neighborhoods.
            Residential neighborhoods, even located far away, can demonstrate strong correlational patterns. 
            This module enables the model to capture these spatial interactions within a single modality, such as the flow of traffic within a city, by learning the spatial structure and dependencies effectively. This approach is crucial for understanding how changes in one location can influence conditions in nearby locations.
            \item Learning temporal dependencies:  This step is motivated by the need to understand how conditions evolve over time, such as temporal dynamics that exhibit periodic patterns, such as within-day traffic flow variations. The temporal attention network  module is designed to identify and prioritize the most relevant past time slices to improve future predictions. By attending to relevant time lags, the model can better anticipate future states by learning from the temporal patterns and historical fluctuations in the data.
        \end{itemize}

    \paragraph{Phase 2 (Inter-Modal Learning Module):} We then discuss our unique inter-modal learning phase. 
    We break down the inter-modal dependencies into local microscopic and global macroscopic levels. 
    a fusion block containing a global-fusion module and a local-fusion module to capture the inter-modal dependencies of spatial units. We then conclude with a multiple perspective interaction module to selectively combine the node embeddings.
 
    \begin{itemize}
            \item   Global fusion and local fusion strategies: The motivation behind employing both global and local fusion strategies lies in the unique necessity to capture spatial dependencies at different spatial scales and resolutions across multiple modalities. The global fusion strategy aims to understand broad, coarse-grained spatial relationships, which is essential for modeling interactions at a macro urban level, such as city-wide traffic patterns. Conversely, the local fusion strategy targets fine-grained local spatial dependencies, enabling the model to capture detailed interactions at a micro level, like specific street corners or neighborhoods. This dual-fusion strategy ensures comprehensive modeling of spatial dependencies, enhancing the model's ability to predict spatial phenomena accurately.
            \item Multiple perspective interactions: This module is motivated by the need to effectively integrate and enhance features from different modalities. By allowing for the fusion of embeddings, this module enriches the node embeddings of spatial unit, ensuring that the representation of each node benefits from a holistic view that incorporates information from various information sources processed in the two sub-modules above. This sub-module improves the model's ability to understand complex spatial and temporal dynamics learned above by leveraging the unique information from different modalities.
     \end{itemize}
     \paragraph{Phase 3 (Prediction Decoder Module):} The final prediction component forecasts future states based on learned spatial and temporal patterns in the previous two steps. By acting as a decoder that uses the framework's learned embeddings to predict the next time step's OD flow matrix, this module aims to translate the complex, multimodal spatiotemporal patterns into OD flow predictions for a focal transportation mode.

\begin{figure*}[!ht]	
		\centering
		\includegraphics[width=\linewidth]{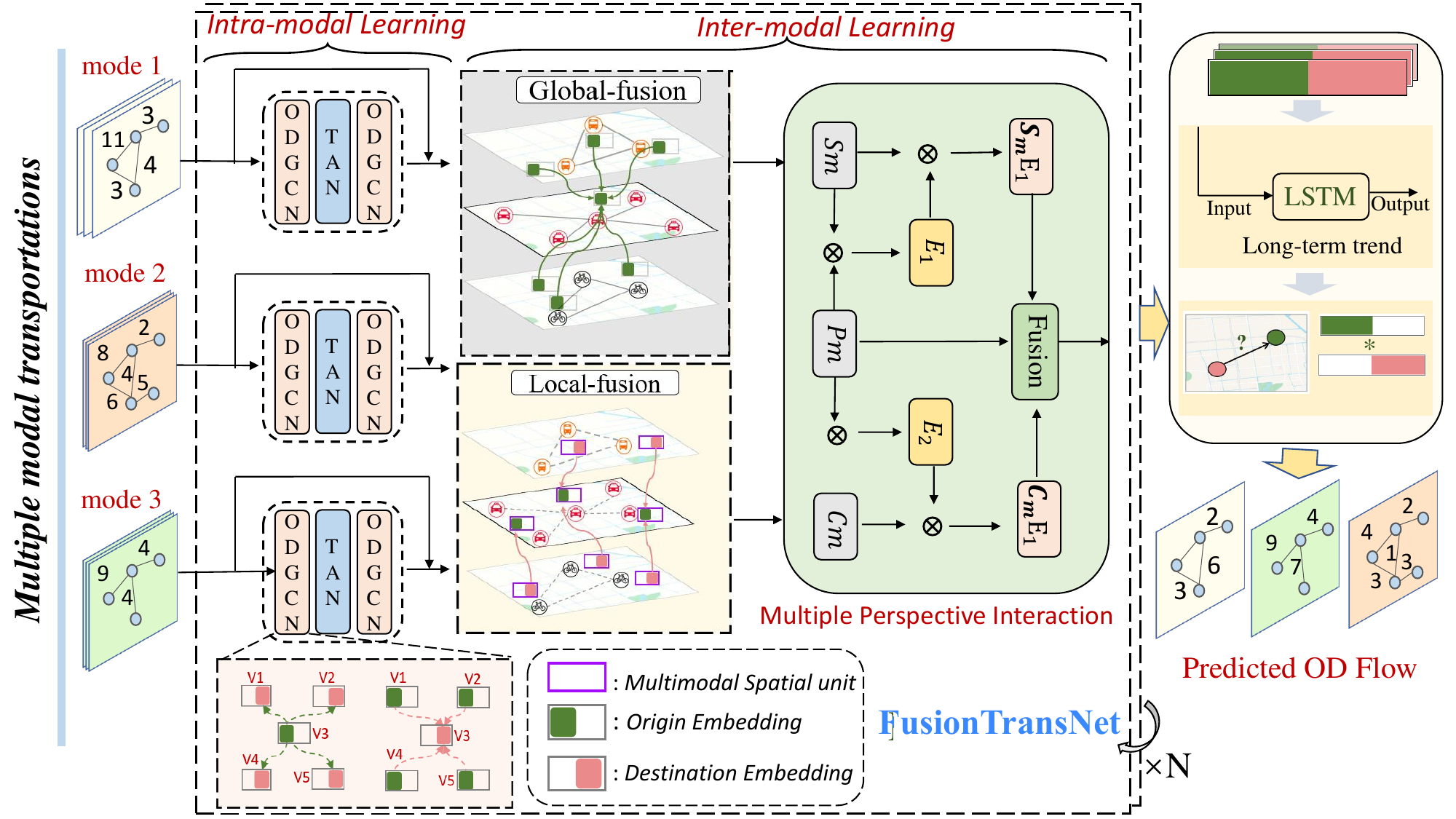}
		\caption{Architecture of \textsf{FusionTransNet}, including the intra-modal learning phase, the inter-modal learning phase, and a final prediction decoder to predict the OD flow. }
		\label{fig:framework}
    \end{figure*}

We next introduce the mathematical formulations of each layer. The detailed architecture is presented in Figure~\ref{fig:framework}. 
    
 \subsection{Phase  1:  Intra-modal Learning in Urban Transportation Systems}

 \label{subsub_intra}
	Intra-modal learning refers to using data on the focal mode to predict its OD flows. 
	That is, we predict $\mathbf{M}_m^{t+1}$ for focal mode $m$ only with the spatiotemporal graphs in the past $L$ time periods for the focal mode $m$ ($\{\mathcal{G}_m^t\}_{t-L}^{t-1}$). 
	Because of the graph-structured information and the spatiotemporal dependencies, we leverage the prediction ability of Graph Convolutional Networks (GCNs) in handling complex inter-connected structures to leverage the strength of the OD matrix of the focal modal $m$. 
	The key to deploying GCNs is to construct the graph structure such that we can improve the predictive power of the focal node. 
    We perform two types of graph convolutions to learn two different embeddings for the same spatial unit, so that we can capture its different roles when serving as roles of origins and destinations in a traffic flow prediction task. 
    

    \paragraph{\textbf{OD-Adaptive-GCN through Interconnected spatiotemporal Graphs.}}
                                                                                       
    To fully capture the interdependencies within a transportation network, we  build several interconnected spatiotemporal graphs that synergize both spatial and temporal aspects of data. By distinguishing between nodes serving as origins and those serving as destinations, we achieve a nuanced view of transportation dynamics. Such granularity is particularly essential considering the inherent asymmetry in transportation data, where departure patterns from an origin can differ considerably from arrival patterns at a destination. Extending the graph-based approach in \citet{wang2018graph}, we build multiple spatiotemporal graphs for each mode, and carefully refine the connections of the inter-modal relationships later in Section~\ref{subsub_inter}, using local-fusion and local-fusion strategies, as well as the \textsf{ModeDistinctNet}.

Traditional spatial networks tend to represent transport hubs (nodes) and their associated flows (edges) as static entities. Introducing temporal granularity allows us to track and analyze dynamic shifts in traffic patterns over time, significantly augmenting prediction capabilities. Our approach involves the construction of two distinct spatiotemporal graphs. The first is grounded on the strength of flow between origins and destinations at specific time points. The second leverages GCNs, particularly node adaptive parameter learning, to discern unique node-specific patterns and recognize spatial correlations. 
    We first use the dynamic traffic flows from origins to destinations at each time step to represent time-dependent relationships between spatial units $i$ and $j$. 
    For each origin spatial unit $i$, we calculate normalized traffic flows to other destinations according to the OD flow matrix at $t$ time-step, relative to the total traffic flow originating from node $i$: 
	\[\textit{Traffic flow normalization:} \quad \mathbf{A}^t_m[i,j]=\frac{\mathbf{M}^t_m[i, j]}{\sum_{k\in \mathcal{V}_m} \mathbf{M}_m^t[i,k]},\] 
    where $\mathbf{M}^{t}_m[i, j]$ corresponds to the traffic flow from spatial unit $i$ to $j$ for mode $m$ at time $t$.

    We next enhance the traditional GCN with node adaptive parameter learning and data-adaptive graph generation to learn node-specific patterns. 
    This adaptive graph generation learns an adjacency matrix end-to-end through stochastic gradient descent, which is effective in discovering hidden spatial dependencies within the mode. 
    That is, to adaptively learn complex spatial dependencies between each pair of spatial units, we use the following function to learn \emph{adaptive OD-graph} $\mathbf{A}^{{S}}_m$ for each mode $m$: 
	\begin{equation}
    \textit{Adaptive OD-graph:}  \quad   \mathbf{A}^{{S}}_m=\operatorname{SoftMax}\left(\operatorname{ReLU}\left(\mathbf{E}_{o} \mathbf{E}_{d}^{T}\right)\right), 
	\end{equation}
	where $\mathbf{E}_o, \mathbf{E}_d \in \mathbb{R}^{N \times d_e}$ are the two randomly initialized  node embedding dictionaries with learnable matrices; 
     $d_e$ denotes the dimension of this initialized node embedding.  
    These adaptive node embedding $\mathbf{E}_o$ and $\mathbf{E}_d$ can capture the inter-dependencies among spatial units that serve as the origins and destinations separately.  
    Both ReLU (Rectified Linear Unit)  and SoftMax are nonlinear activation functions. 
    We use ReLU activation to eliminate weak connections and the SoftMax function to normalize the row sum of adaptive OD-graph to one.

    The incorporation of adaptive node embeddings tailored for origin and destination, separately, allows the model to better cater to the nuances of traffic flow in the urban areas. This adaptability is crucial for complex transportation networks where spatial units can have varying traffic flow patterns when they act as origins or destinations. 
    Specifically, given an input at time step $t$ of mode $m$ is denoted as $\mathbf{H}_{m,t} \in \mathbb{R}^{N_m \times d_c}$, where $d_c$ is the dimension of latent embeddings . We then decompose $\mathbf{H}_{m, t}$ into two vectors separately for the origins and destination. 
    This step results in an origin embedding vector $\mathbf{H}^O_{m,t}$ and a destination embedding vector $\mathbf{H}^D_{m,t}$ for mode $m$ at time $t$, respectively. 
	In this case, we have: 
        \begin{equation}
            \mathbf{H}^O_{m,t} =\mathbf{H}_{m,t}[:, :\frac{d_c}{2}] \quad \text{and} \quad \mathbf{H}^D_{m,t} =\mathbf{H}_{m,t}[:,  \frac{d_c}{2}:], 
        \end{equation}
    where $\mathbf{H}^O_{m,t}, \mathbf{H}^D_{m,t}\in \mathbb{R}^{N_m \times \frac{d_c}{2}}$ can represent the flow generation and attractive features of $N_m$ nodes for mode $m$, each with dimension $\frac{d_c}{2}$.  
    
    We then perform convolutions on the normalized OD flow matrix $\mathbf{A}_m^{t}$ and adaptive OD matrix $\mathbf{A}^{S}_m$ to capture flow patterns from origins to destinations. 
    The origin embedding $\mathbf{Z}^O_{m,t} \in \mathbb{R}^{N_m \times \frac{d_e}{2}}$  for mode $m$ at time $t$ can be calculated as:     
	\begin{equation}
    \textit{Intra-modal origin embedding:}  \quad   \mathbf{Z}^O_{m,t}=\sum_{\mathbf{A}_{c} \in \{\mathbf{A}_m^{t}, \mathbf{A}^{S}_m\}} \operatorname{ReLU}\left( (\mathbf{A}_{c})^T \mathbf{H}^O_{m,t} \mathbf{W}_{c}^{O}+b_{c}^{O}\right)
	\end{equation}
    where $\mathbf{W}_{c}^{O} \in \mathbb{R}^{\frac{d_c}{2} \times \frac{d_e}{2}}$ and $b_{c}^{O} \in \mathbb{R}^{N_m}$ are the learnable parameters. 
	
	Similarly, to learn the destination embedding  $\mathbf{Z}^D_{m,t} \in \mathbb{R}^{N_m \times d_e/2}$ of mode $m$ at time $t$, we transpose the adjacency matrices of the two directed graph structures, and we then perform the convolution operation again:
	\begin{equation}
		    \textit{Intra-modal destination embedding:}  \quad \mathbf{Z}^D_{m,t}=\sum_{\mathbf{A}_{c} \in\{\mathbf{A}_m^{t}, \mathbf{A}^{S}_m\}} \operatorname{ReLU}\left(\left(\mathbf{A}_{c}\right)^T \mathbf{H}^D_{m,t} \mathbf{W}_{c}^{D}+b_{c}^{D}\right)
	\end{equation}
	where $W_{c}^{D} \in \mathbb{R}^{d_c/2 \times d_e/2}$ and $b_{c}^{D} \in \mathbb{R}^{N_m \times d_e/2}$ are learnable parameters. 
 
 Finally, we stack the origin embedding vector $\mathbf{Z}^O_{m,t}$ and the destination embedding vector $\mathbf{Z}^D_{m,t}$ of mode $m$ at time $t$  along the rrigin and destination dimension, and the output is denoted as: 
 \[		    \textit{Intra-modal spatial unit embedding:}  \quad \mathbf{Z}_{m,t} = [\mathbf{Z}^O_{m,t}  || \mathbf{Z}^D_{m,t}],\] 
 where $[\cdot || \cdot] $ is an concatenation operator. 
 
	\paragraph{\textbf{Temporal Attention Network.}} 
 We further introduce a temporal attention module to model dynamic temporal correlations. The utilization of a temporal attention mechanism introduces the capacity to weigh the importance of different historical time steps. 
 By attending to different time scales, the model can capture both short-term, long-term, and sustained temporal correlations. This is particularly important for understanding traffic patterns which can have cyclical (daily or weekly) trends, seasonality, or react to sporadic events.
Given the input sequence $\mathbf{Z}_m \in \mathbb{R}^{L \times N \times d_e} =\left[Z_{m,t-L},...,Z_{m,t-1},Z_{m,t}\right]$,  the temporal attention module can be defined as follows:

 \begin{equation}
		\begin{aligned}
	\textit{Temporal dependency:} \quad &\mathbf{E}_e= \mathbf{W}^0_e \cdot \sigma\left(\left(\mathbf{Z}_m^\top \mathbf{W}_e^1\right) \mathbf{W}_e^2\left(\mathbf{Z}_m \mathbf{W}_e^3\right)+b_e\right) , \\
    \textit{Normalized temporal dependency:} \quad & \mathbf{E}^{\prime}\left[i, j\right]  = \text{SoftMax} (\mathbf{E}_e\left[i, k\right]) 
	\textit{Temporal attention:} \quad & \mathbf{P}_m  =  \mathbf{E}^{\prime}\mathbf{Z}_m, 
 \label{eq:intra_output}
		\end{aligned}
	\end{equation}
 where $\mathbf{W}^0_e \in \mathbb{R}^{L \times L}$ and $ b_e \in \mathbb{R}^L$ ($L$ denotes the time length of the input); $\mathbf{W}_e^1 \in \mathbb{R}^{N_m}$, $\mathbf{W}_e^2 \in \mathbb{R}^{d_e \times N_m}$, and $\mathbf{W}_e^3 \in \mathbb{R}^{d_e}$ are learnable parameters. The value of an element $\mathbf{E}_e\left[i, j\right]$ in $\mathbf{E}_e \in \mathbb{R}^{L \times L}$ represents the strength of dependencies between time $i$ and $j$. 
 Then $\mathbf{E}_e$ is normalized by the SoftMax function and used to re-weight different time steps of the input. 
 Finally, we can get the output vector $\mathbf{P}_m \in \mathbb{R}^{N_m\times d_p}$, where $d_p$=$L \times d_e$.

    \subsection{Phase 2: Inter-modal Learning: Dissecting Cross-Transportation-Mode Dynamics at Macro and Micro Levels}
     \label{subsub_inter}
    Multi-modal travel is common in modern transportation systems, where users often switch between different modes of transport within a single journey. Modeling this system requires an understanding of how spatiotemporal patterns in one mode (e.g., subway) might affect another (e.g., bus), as motivated in Figure~\ref{fig:pattern}.  The goal of inter-modal learning is to enhance the spatiotemporal representation of data from each transportation  modal and improve model prediction performance based on the information complementarity of different transportation modals.

     Our unique inter-model learning phase consists of three components (as shown in Figure~\ref{fig:skeleton}): the global fusion, local fusion, and multiple perspective interaction module. 
     To start with, the global fusion strategy provides a high-level view by aggregating patterns across different transportation modalities. By learning patterns from global aggregations, it allows us to capture broader, system-level dynamics.
     Then, we propose a local-fusion strategy to capture fine-grained spatial dependencies and detailed interactions among multiple modals at a micro level.
     Finally, we design a multiple perspective interaction module to integrate the two learned features from the global and the local strategies. 
	
    \paragraph{\textbf{Global-Fusion Strategy at Macroscopic Urban System Level.}}  

    This strategy focuses on capturing higher-level patterns and dynamics across different transportation modes. Essentially, this strategy aims to model the traffic flows in the entire urban transportation system. By using global aggregations, the model attends to patterns that are recurrent across different modes. 
    We use an attention mechanism to distill predictive input traffic flow patterns and to capture similarity among origin and destination across modalities~\citep{vaswani2017attention}. 

     We use the following examples to motivate global-fusion strategy in inter-modal learning. 
    \begin{example}[Impact of rush hours on urban mobility]
    Consider the dynamics of a typical weekday in a bustling city. During morning and evening rush hours, the entire transportation infrastructure—buses, trains, and roads—experiences significantly increased usage. A global fusion strategy plays a crucial role here by simultaneously analyzing data across these modalities to identify overarching daily patterns. For instance, it might reveal that train delays during peak hours lead to a noticeable increase in road traffic as commuters switch to alternative modes of transportation, such as taxis.
    \end{example}

    \begin{example}[Response to citywide events]
    Imagine a scenario where a major event disrupts the city's transport ecosystem. Such an event might cause buses to reroute, subways to become overcrowded, road traffic to slow down, and an uptick in bicycle usage as commuters seek alternative routes and modes of transport. Employing a global fusion strategy allows for the integration of data from all affected transportation modes, offering a comprehensive view of the event's impact on the city's mobility network.  
        \end{example}

    \paragraph{Inter-modal Attention for Predictive Inter-modal Patterns:}
    We decompose two input vectors $\mathbf{P}_m^O$ and $\mathbf{P}_m^D$ from $\mathbf{P}_m$ (calculated from Equation~\eqref{eq:intra_output})  as the origin and the destination embedding vectors, and we then use attention mechanism to extract similar flow patterns among origins or destinations. 
    The matrix $\mathcal{E}_{m,n}^{O}, \mathcal{E}_{m,n}^{D} \in \mathbb{R}^{N_m \times N_n}$ measures the similarity among the origins (destinations) of modal $m$ and the origins (destinations) of modal $n$: 
    \begin{equation}
    \begin{split}\textit{Inter-modal global-fusion attention:} \quad
    \begin{cases}    
    & \mathcal{E}_{m,n}^{O}\left[i,j\right]  = \text{SoftMax}_n (LeakyReLU\left(\left[\mathbf{P}^O_{i,m}\mathbf{W}_g\|\mathbf{P}^O_{j,n} \mathbf{W}_g\right]\right) \\ 
  \quad & \mathcal{E}_{m,n}^{D}\left[i,j\right]  = \text{SoftMax}_n \left( LeakyReLU \left(\left[\mathbf{P}^D_{i,m}\mathbf{W}_g\|\mathbf{P}^D_{j,n} \mathbf{W}_g\right]\right)\right) 
    \end{cases}
    \end{split}
 \label{eq:global_attention}	
 \end{equation}
	where $\mathbf{P}^O_{i,m}$ is the embedding vectors of the origin $v_i$ of modal $m$, and $\mathbf{P}^O_{j,n}$ means the origin $v_j$ of modal $n$, respectively. $LeakyReLU(\cdot)$ denotes the Leaky Rectified Linear Unit function, and $\mathbf{W}_g$ is a learnable matrix. 
 
    \paragraph{ModeDistinctNet for Cross-Modal Distinct Patterns:}
    Modeling the dynamics of a multimodal transportation system requires careful consideration of both the similarities and differences across various modes. The local attention weights, as described in Equation~\eqref{eq:global_attention}, are adept at learning the similarities among different nodes. However, this focus on similarities may inadvertently overlook the unique characteristics and idiosyncrasies of each node, potentially leading to negative transfer by overfitting to historical spurious correlations. 
    To mitigate the risk of negative transfer and to fully harness the diversity of the urban system, it is crucial to model the distributional differences between modes. This idea stems from the recognition that each mode of transportation—be it buses, trains, or cars—possesses its own set of flow patterns. For example, the implications of a delay in the subway system are likely to differ significantly from those in the taxi system. 

    In response to this challenge, we designed \textsf{ModeDistinctNet} sub-module to capture the intrinsic differences in inflow and outflow features between a focal node \(i\) and another node \(j\), across different transportation modes. By focusing on the distinct inflow/outflow characteristics of origins and destinations, and how these characteristics differ from one node to another, we can obtain a more accurate representation of different transportation modes. This approach ensures that the distinct nature of each mode is preserved and that the aggregation of information across modes is conducted in a manner that preserves or enhances the model's accuracy. 
    Therefore, the following operation is designed for avoiding specific points of potential negative transfer by highlighting where the flow characteristics diverge significantly.
    \begin{equation}
    \begin{split}
    \textit{Inter-modal embeddings from {ModeDistinctNet}:} \quad
    \begin{cases}   \mathbf{c}^O_{i, m}&=\sum_{{n} \in \mathcal{T}_i}\sum_{j \in N_n} \left(\mathcal{E}_{m,n}^{O}\left[i,j\right] \left|\mathbf{P}^O_{i,m}\mathbf{W}^O_g-\mathbf{P}^O_{j,n}\mathbf{W}^O_g \right|\right), \\
  	\mathbf{c}^D_{i, m}&                                                                                                                                                                                                                  =\sum_{{n} \in \mathcal{T}_i}\sum_{j \in N_n} \left(\mathcal{E}_{m,n}^{D}\left[i,j\right] \left|\mathbf{P}^D_{i,m}\mathbf{W}^D_g-\mathbf{P}^D_{j,n}\mathbf{W}^D_g \right|\right). 
   \end{cases}
  \end{split}
	\end{equation}
    Here, $\left|\mathbf{P}^O_{i,m}\mathbf{W}^O_g-\mathbf{P}^O_{j,n}\mathbf{W}^O_g \right|$ represents the absolute difference in outflow characteristics, weighted by a global learnable parameter $\mathbf{W}_g^O$, which quantifies the disparities between origins. 
    Similar calculation applies for destinations. 
    The double summation traverses all modes in the spatial units $i$ and their node set.

    Finally, we concatenate $\mathbf{c}^O_{m}$ and $\mathbf{c}^D_{m}$ along with the feature dimension to obtain a globally-aggregated global-fusion vector: 
    \[ \textit{Global inter-modal embedding:} \quad \mathbf{c}_m =  [\mathbf{c}^O_{m} \| \mathbf{c}^D_{m}], \]
    where $\mathbf{c}_m  \in \mathbb{R}^{N_m\times d_f}$.

	\paragraph{\textbf{Local-Fusion Strategy at Microscopic Multi-modal Spatial Unit Level.}} 

The local fusion strategy is designed to explore spatial correlations within and between origins and destinations across different transportation modes. It addresses the detailed aspects of passenger transfers and the use of different modes, emphasizing the direct interactions that affect the efficiency of the entire transportation network.
This local-fusion approach contrasts with the global fusion strategy, which broadens the perspective to city-wide patterns and trends across transportation modes. Instead, the local fusion strategy narrows down to examine the direct relationships at a more detailed level, especially within the context of multimodal transit stations, to capture the immediate dynamics. To illustrate the importance of this approach, consider the following example.
    \begin{example}[Necessity of Local Fusion]
Consider the scenario where a bus line malfunction causes delays. The multiple perspective interaction module discerns the interconnected impacts across modes—predicting an uptick in taxi usage for longer commutes and a rise in bicycle trips for shorter distances. This insight is derived from an examination of how disruptions in one mode propagate effects throughout the network, underscoring the module's capability to capture and analyze complex inter-modal interactions beyond simple aggregate or isolated observations.  
    \end{example}

    The local fusion strategy utilizes an attention mechanism to assess how similar different transportation modes are within a specific area. By examining the inflow and outflow data for each mode at points where multiple modes intersect (multimodal units), the strategy aims to better understand local traffic patterns. This enhanced understanding helps the model accurately reflect the unique dynamics at these nodes.

    To quantify the similarity, we look at the inflow and outflow embeddings for origins and destinations across modes. Specifically, we calculate the similarity between the inflow of origins and the outflow of destinations for different modes, represented by $\mathbf{E}_{i}^O\in \mathbb{R}^{\left|\mathcal{T}_i\right| \times \left|\mathcal{T}_i\right|}$, at a multimodal spatial unit $v_i$. This calculation is performed using an attention mechanism to weigh the importance or relevance of the data points: 
 \begin{equation}
   \textit{Inter-modal local-fusion attention:} 
    \begin{cases}    
& \mathbf{E}_{i}^O\left[m,n\right]= \text{SoftMax}_n \big(  LeakyReLU \left(\left[\mathbf{P}^O_{i,m}\mathbf{W}_l^O\|\mathbf{P}^O_{i,n}\mathbf{W}_l^O\right]\right)\big) \\
& \mathbf{E}_{i}^D\left[m,n\right]= \text{SoftMax}_n \big(  LeakyReLU \left(\left[\mathbf{P}^D_{i,m}\mathbf{W}_l^D\|\mathbf{P}^D_{i,n}\mathbf{W}_l^D\right]\right)\big)
         \label{eq:local_attention}
    \end{cases}
\end{equation}
     where $\mathbf{W}_{l}^O$ and $\mathbf{W}_l^D$ are learnable matrix; 
	$\mathcal{T}_i$ is the set of traffic modals at the multi-modal spatial unit $i$. 
    Specifically, inter-modal attention is calculated by comparing the inflow and outflow embeddings of different modes at a multimodal unit. This process effectively highlights the relationships between modes based on their traffic patterns at these critical points.
    
    The difference between the inter-modal attention in global fusion (Equation~\eqref{eq:global_attention}) and local fusion (Equation~\eqref{eq:local_attention}) lies in the scope of the attention network. 
    The global-fusion component considers the relationships of all origins across all modules and the nodes therein, hence the goal is to find similar units across the entire urban system. 
    By contrast, the local-fusion here  considers cross mode relationships within the same spatial units, hence the goal is to find a similar mode at a local spatial unit level. 
    
    We then use \textsf{ModeDistinctNet}, as mentioned above, to aggregate information for a multi-modal spatial unit $i$: 
	\begin{equation}
 \begin{split}
     \textit{Inter-modal local-fusion embedding:} \quad
     \begin{cases}
    \mathbf{S}^O_{i, m}=\sum_{n \in \tau_i} \big(\mathbf{E}_{i}^O\left[m,n\right] \left| \mathbf{P}^{O}_{i, m}\mathbf{W}_l^O-\mathbf{P}^D_{i,n}\mathbf{W}_l^O \right|\big)\\
    \mathbf{S}^D_{i, m}=\sum_{n \in \tau_i} \big(\mathbf{E}_{i}^D\left[m,n\right] \left| \mathbf{P}^{D}_{i, m}\mathbf{W}_l^D-\mathbf{P}^O_{i,n}\mathbf{W}_l^D \right|\big)
 \end{cases}
 \end{split}
	\end{equation}
	where $\mathbf{W}_{l}^O$ and  $\mathbf{W}_{l}^D$ are two  learnable matrices.
    $\mathbf{S}^O_{i, m}$ and  $\mathbf{S}^D_{i, m}$ extract similar features between the inflow (outflow) of modal $m$ and the outflow (inflow) of the other modals at the multi-modal spatial unit $i$.
    
    Finally, we concatenate these two vectors to obtain locally-aggregated embeddings for multi-modal spatial units $\mathbf{S} \in \mathbb{R}^{N_m\times d_s}$: 
    \[\textit{Inter-modal spatial-unit embedding:} \quad \mathbf{S}^O_{m} =  [\mathbf{S}^O_{m}||\mathbf{S}^D_{m} ].  \]

 \paragraph{\textbf{Multiple Perspective Interaction Module.}} 

 
This module integrates insights from both global and local scales to enrich the representation of each node within the multi-modal network. 
The module's primary objective is to synthesize insights from global trends, such as increased transportation demand during sports events, with localized phenomena, like weather impacts on bus and bike usage. This module combines information from macro and micro level observations while also enhances the individual representations by leveraging the synergy between global and local features.

Leveraging attention mechanisms, the module extracts and integrates pertinent information from both globally and locally aggregated data. This process involves the construction of attention matrices to distill relevant features, which are then seamlessly incorporated into the original node representations. This fusion mechanism enriches the model's ability to depict the multifaceted nature of urban transportation systems.

To formalize the integration, we employ two attention matrices $\mathbf{E}_1 \in \mathbb{R}^{d_p\times d_c}$ and $\mathbf{E}_2 \in \mathbb{R}^{d_p\times d_s}$, designed to selectively extract information from the global ($\mathcal{C}_m$) and local ($\mathcal{S}_m$) contexts, respectively. 
This selective emphasis is achieved through: 
    \begin{equation}
		\begin{aligned}
			\textit{Intra-modal and \textbf{global} inter-model embeddings:} \quad & \mathbf{E}_g=\sigma\left(\left(\mathbf{P}_m \mathbf{W}_p^1\right)^\top\left(\mathbf{c}_m \mathbf{W}_p^2 \right)\right), \\
			\textit{Intra-modal and \textbf{local} inter-model embeddings:}  \quad & \mathbf{E}_l=\sigma\left(\left(\mathbf{P}_m \mathbf{W}_p^3\right)^\top\left(\mathbf{S}_m \mathbf{W}_p^4\right)\right), 
		\end{aligned}
    \label{eq:multiple_perspec_1}
	\end{equation}
    where $\sigma(\cdot)$ is the Sigmoid activation function. $\mathbf{W}_p^1$, $\mathbf{W}_p^2$, $\mathbf{W}_p^3$, and $\mathbf{W}_p^4$ are all learnable matrices.  The integration of these focused information into node representations is formulated as:
     \begin{equation}
		\textit{Final embedding:} \quad \mathbf{U}_m = \mathbf{W}_f^1 \odot \mathbf{P}_m+\mathbf{W}_f^2 \odot\left(\mathbf{c}_m\mathbf{E}_g^{\top} \right)+\mathbf{W}_f^3 \odot\left(\mathbf{S}_m\mathbf{E}_l^{\top} \right), 
    \label{eq:multiple_perspec_2}
	\end{equation}
    where each term represents the contribution of raw node features, globally-aggregated, and locally-aggregated feature vectors, modulated by learnable weight matrices ($\mathbf{W}_f^1$, $\mathbf{W}_f^2$, and $\mathbf{W}_f^3$) and combined through the Hadamard product ($\odot$).

	\subsection{Phase 3:  Prediction Decoder  Module}

Following the integration of insights through fusion techniques, the Prediction Module's primary goal is to accurately forecast future OD flow dynamics within the transportation network. Utilizing LSTM layers, a type of recurrent neural network architecture, this module can effectively learn the long-term temporal patterns and dependencies. This capability is important for predicting the flow between various points—origins and destinations—over future time steps.  

Based on the refined origin and destination embeddings derived from prior global-fusion and local-fusion components, the module predicts the OD flow for each modality, denoted as  $m$, from origin $i$ to destination $j$  at the subsequent time step. 
We decompose $\mathbf{U}_m$ into origin embedding $\mathbf{U}_m^O$ and destination embedding vector $\mathbf{U}_m^D$. 
     \begin{equation}	\hat{\mathbf{M}}^{t+1}_m[i,j]=\left(\mathbf{W}_m \mathbf{U}^O_{i,m}\right)\left(\mathbf{U}^D_{j,m}\right)^T, 
     \label{eq:pred}
	\end{equation}
 
 where \(\mathbf{W}_m \in \mathbb{R}^{N_m \times N_m} \) represents a matrix of learnable parameters, while \(\mathbf{U}^O_{i,m}\) and \(\mathbf{U}^D_{j,m}\) signify the embedding vectors for the origin $i$ of modal $m$ and destination nodes $j$ of modal $m$, respectively. Eqn.~\eqref{eq:pred} forms the basis for the module's predictive output, with accuracy being assessed via the $L_2$ loss function:
	\begin{equation}
		\mathcal{L}_m= \left\|\mathbf{M}_{m}^{t+1}-\hat{\mathbf{M}}^{t+1}_{m}\right\|
	\end{equation}
	where $\mathbf{M}_m^{t+1}$ represents the ground truth OD matrix of modal $m$ at the time step t+1.

To further enhance the model's performance across different transportation modes, a weighted multimodal loss function is introduced. This new approach ensures that the model's learning is not disproportionately influenced by data from any single modality. It achieves this by dynamically weighting the loss from each modality, taking into account variations in data magnitude: 
	\begin{equation}
		\mathcal{L}=\sum_{m=1}^{K}\frac{\eta_m}{\mu_m}\mathcal{L}_m, 
	\end{equation}
In this equation, \(\eta_m\) serves as a hyperparameter for adjusting the weight of the loss from modality \(m\), while \(\mu_m\) acts as a balancing factor, computed based on the ground truth OD matrix \( \mathbf{M}^t_m\). 
    \begin{equation}
		\mu_m=\frac{1}{N_m\times N_m}\sum_{i=1}^{N_m}\sum_{j=1}^{N_m} \mathbf{M}^t_m[i,j], 
	\end{equation}
 where $N_m$ is the number of spatial units in mode $m$.

 This predictive capability allows the model to not only predict general trends but also respond to specific events, such as a subway malfunction, by forecasting consequent changes in demand for other transportation modes (e.g., buses and taxis). Such prediction enables proactive measures by city officials or transportation agencies, enhancing the adaptability and efficiency of urban transportation systems. 
By synthesizing large-scale patterns through global fusion, capturing detailed patterns via local fusion, and refining information through  multiple perspective interaction module, the prediction module makes the final prediction on OD flows. 

	\section{Numerical Evaluations}
 The experimental section is structured to provide an evaluation of the methodologies and results obtained during this research. 
 In Section~\ref{subsec:datasets}, a  description of the datasets employed and the experimental procedures undertaken is provided. 
 Section~\ref{subsec:performance_comparisons} provides a comparative analysis wherein the performance metrics of the proposed model are compared against existing benchmarks. 
 In Section~\ref{subsec:ablation}, an ablation study is presented to ascertain the integral value of each component of the model. 
 Section~\ref{subsec:interpretations} delves into an interpretation of both the global and local fusion strategies, elucidating their individual contributions and the synergistic effects when integrated. The section concludes with an analysis of the interplay between these two strategies, emphasizing their combined efficacy in the context of the overall model. 
 
	\subsection{Datasets and Experiment Settings} 
 \label{subsec:datasets}
     The datasets used in this study originate from transportation agencies in two urban centers: Shenzhen and New York. 
     With a population of 17.56 million in 2020, Shenzhen is the third most populous city by urban population in China after Shanghai and Beijing.\footnote{The city is a leading global technology hub. In the media Shenzhen is sometimes called China's Silicon Valley.  The residents of Shenzhen are made up of immigrants from all over China, possessing the youngest population structure in the country and an anti-discrimination urban culture.}
     The dataset from Shenzhen encompasses three distinct transportation modalities, namely taxi, shared bikes, and buses. In comparison, the New York dataset focuses on two modalities, which are taxis and shared bikes. Taxi data is derived from GPS-based systems, capturing in detail the trajectories of the taxis and all trips. This data provides insight into routes and occupancy patterns. The shared-bike dataset, on the other hand, is constructed from order records harvested from bike-sharing applications, offering a snapshot of user demands and peak usage times. The Shenzhen bus dataset, sourced from the Integrated Circuit smart ticketing system, offers a unique vantage point, mapping out bus routes, occupancy, and frequency. A detailed specification of both datasets, including aspects like the number of entries, transportation modalities, and time frames, is presented in Table~\ref{tab:dataset}.  
     

	\begin{table}[!htp]
	\caption{Data descriptions with multiple traffic modals.}
 	\label{tab:dataset} 
    \centering 
	\begin{tabular}{c|c|c}
		\midrule
		City                                & Shenzhen                                 & New York                                \\ \midrule
		\multicolumn{1}{l|}{Transportation} & Taxi/Bus/Bike & Taxi/Bike                              \\
		Time range                          & \multicolumn{1}{l}{7/1/2017-7/30/2017}   & \multicolumn{1}{l}{7/1/2017-7/30/2017} \\
		Time span                           & 1 hour                                   & 1 hour                                 \\
		Grid                                & 491                                    & 225                                  \\
		Stations                            & 482/451/484                         & 75/46                                  \\
		Node feature                        & Inflow/Outflow                           & Inflow/Outflow/POI                     \\ \midrule
	\end{tabular}

\end{table}

\begin{table}[!htp]
    \centering
    \begin{tabular}{c|c|c|c|c|c|c|c|c|c|c|c}
        \toprule
        & \multicolumn{6}{c|}{Shenzhen} & \multicolumn{4}{c|}{New York} \\

        \cline{2-11}
        & Taxi & Bus & Taxi & Bus & Bike & Bike & Taxi & Taxi & Bike & Bike & \\
        \cline{2-12}
        City & Task & Task & Task & Task & Task & Task & Task & Task & Task & Task & \\
        \cline{2-12}
        Metrics & MAE & RMSE & MAE & RMSE & MAE & RMSE & MAE & RMSE & MAE & RMSE & \\
        \midrule
        HA & 0.1949 & 0.2043 & 1.6521 & 2.5687 & 0.3654 & 2.0166 & 5.0045 & 7.9359 & 0.9412 & 3.1277 & \\
        LSTM & 0.1865 & 0.1923 & 1.5348 & 2.5386 & 0.2487 & 1.8605 & 3.5432 & 5.9803 & 0.8364 & 2.9402 & \\
        ConvLSTM & 0.1812 & 0.1832 & 1.5199 & 2.4454 & 0.2390 & 1.8724 & 3.9509 & 5.2681 & 0.8208 & 2.9348 & \\
        GCRN & 0.1632 & 0.1645 & 1.3103 & 2.2139 & 0.2245 & 1.7323 & 2.9456 & 4.7321 & 0.7134 & 2.8676 & \\
        CSTN & 0.1562 & 0.1512 & 1.2132 & 2.1034 & 0.2037 & 1.6876 & 2.8123 & 4.3492 & 0.6465 & 2.8140 & \\
        GEML & 0.1512 & 0.1361 & 1.1841 & 1.9723 & 0.1912 & 1.5478 & 2.6991 & 3.9423 & 0.6345 & 2.7123 & \\
        DGDR & 0.1484 & 0.1321 & 1.1855 & 1.9681 & 0.1843 & 1.5594 & 2.7194 & 3.9626 & 0.6278 & 2.6715 & \\
        CMOD & 0.1430 & 0.1272 & 1.1864 & 1.9688 & 0.1761 & 1.4670 & 2.7176 & 3.9634 & 0.6283 & 2.6723 & \\ \midrule 
        \rowcolor{gray!30} \textsf{FusionTransNet} & \textbf{0.1353} & \textbf{0.1217} & \textbf{1.1128} & \textbf{1.8573} & \textbf{0.1635} & \textbf{1.3873} & \textbf{2.5642} & \textbf{3.7662} & \textbf{0.5948} & \textbf{2.5670} & \\
        \bottomrule
    \end{tabular}
    \caption{The prediction performance of the models on the datasets from two cities.}
    \label{tab:performance_comparisons}
\end{table}

 The datasets were divided temporally to ensure that the model was tested on the most recent data. Specifically, 70\% of the data was designated for training, 20\% for validation, and the remaining 10\% for testing. Every OD flow matrix entry corresponds to an hour, ensuring a uniform temporal resolution across the datasets.
 To maintain consistency and facilitate the model's learning process, the Max-Min normalization strategy was employed. This approach standardized all dataset entries to a [0,1] scale. For the evaluation phase, predictions were denormalized to reflect real-world values, ensuring interpretability.
Our model was implemented using PyTorch. Given the variability in real-world datasets, an extensive parameter-tuning process was performed. The AdamW optimizer is used to optimize all the models~\citep{loshchilov2017decoupled}. 
AdamW is a stochastic optimization method that modifies the typical implementation of weight decay in Adam, by decoupling weight decay from the gradient update. 
It has superior convergence properties than the original Adam. 
The model's performance was quantified using the Mean Absolute Error (MAE) and Root Mean Square Error (RMSE) metric. 
MAE and RMSE are defined as follows: 
	\begin{equation}
		\begin{aligned}
			\mathrm{MAE} & = \frac{1}{N_m \times N_m \times T \times |\mathcal{M}|} \sum_{ m \in \mathcal{M}}\sum_{t=1}^T \sum_{j=1}^{N_m} \sum_{i=1}^{N_m}\left| \mathbf{M}^t_m[i,j]-\hat{\mathbf{M}}^t_m[i,j]\right|, \\
			\mathrm{RMSE} & =\sqrt{\frac{ \sum_{ m \in \mathcal{M}}\sum_{t=1}^T\sum_{j=1}^{N_m}\sum_{i=1}^{N_m}\left(\mathbf{M}^t_m[i,j]-\hat{\mathbf{M}}^t_m[i,j]\right)^2}{N_m \times N_m \times T \times |\mathcal{M}|}}. 
		\end{aligned}
	\end{equation}
 Recall that  ${\mathbf{M}}^{t}_m[i,j]$  and $\hat{\mathbf{M}}^{t}_m[i,j]$ are ground-truth and predicted OD flows from the origin $i$ to the destination $j$ of modal $m$ at the next time step. 
 $N_m$ is the number of the origin/destinations for mode $m$. 
 $T$ is the number of total time stamps.

 \subsection{Performance Comparisons}
       \label{subsec:performance_comparisons}
	\subsubsection{Benchmark Methods}
  Our \textsf{FusionTransNet} model was compared against a spectrum of models ranging from simple statistical models to complex deep learning architectures. 
	\begin{itemize}
		\item \textbf{HA}: We use the Historical Average method (HA) to predict the OD demand transactions based on the average of historical value among stations.
		This naive predictor ignores the complexity of OD flow data and the non-linear relationships inherent within it. 
		\item \textbf{LSTM}: The Long Short-Term Memory network is known for its ability to learn long-term temporal patterns from time-series data. We use LSTM to predict the OD flow by modeling the temporal correlations from the historical OD flow matrix.
        Note that LSTM does not incorporate spatial features. 
		\item \textbf{ConvLSTM}~\citep{shi2015convolutional}: The Convolutional LSTM Network is composed of CNN and LSTM to learn spatial and temporal features of the OD flow matrix.
		This framework uses exactly the same dataset as ours. Yet, it can not account for the spatial across-mode dependencies. 
		\item \textbf{GCRN}~\citep{seo2018structured}: The recently proposed Graph Convolutional Recurrent Network (GCRN) uses GCN on geographic networks to learn spatial correlations with RNNs to capture the dynamic patterns.
        Similar to ConvLSTM, this model uses the same datasets as ours. 
        Yet, it improves upon ConvLSTM that can capture the 
		spatiotemporal dependencies by filtering inputs and hidden states passed to a recurrent unit using graph convolution 
		\item \textbf{CSTN}~\citep{liu2019contextualized}: The contextualized spatiotemporal network (CSTN) models local spatial context, temporal evolution context, and global correlation context to predict the future OD flow. 
		The improved spatiotemporal module in this model allows it to capture the spatial correlations more adeptly. 
		\item \textbf{GEML}~\citep{wang2019origin}: GEML uses GCN to aggregate spatial information of geographic and semantic neighbors, and a multi-task learning network is utilized to jointly predict both inflow and outflow for accurate OD demand prediction. 
        The way to use neighboring information offers it an advantage, compared with CSTN. 
		\item \textbf{DGDR}~\citep{dapeng2021dynamic}: DGDR is a joint learning framework that dynamically learns adaptive graph structure for GCN to predict the future OD flow.
		The dynamic nature of DGDR, allowing it to adapt graph structures on-the-fly, offers it an edge, leading to improved performance. 
		\item \textbf{CMOD}~\citep{han2022continuous}: CMOD is a continuous-time dynamic graph representation learning framework and includes a hierarchical message passing module to model the spatial interactions of stations with different granularity.
      CMOD's hierarchical approach to modeling spatial interactions across various granularities results in performance gains over other models. 
	\end{itemize}

     \subsubsection{Performance Comparisons}
    The results of all models under three metrics are shown in Table \ref{tab:performance_comparisons}. For the results of the experiment, we make the following observations. The HA model achieves larger errors because it fails to learn complex nonlinear correlations from historical OD flow data. The performance of LSTM is poorer than SCTN since it only captures temporal patterns and ignores spatial properties among stations, this shows the model which only cares about the time-series feature is not suitable for the OD flow prediction problem. ConvLSTM can capture spatial dependencies using CNN, thus, its prediction performance is better. Further, CSTN models local spatial context and global correlation context using CNN, which integrates external auxiliary features for OD flow prediction. Thus, CSTN learns more information and achieves lower errors than LSTM and ConvLSTM. 
    
    GCRN and GEML achieve better performance than CSTN due to the strong predictive ability of GCN to handle graph-structure traffic data. GEML constructs two static road network structures to model spatial correlation among regions, however, this may not accurately represent complex spatial dependencies. DGDR can dynamically learn graph structures instead of predefined graph structures, so it helps the model capture more complex spatial dependencies and achieves lower errors. CMOD achieves better performance than DGDR because it models the spatial interactions of stations with different granularity. 

    Our proposed \textsf{FusionTransNet} achieves the best performance on various metrics since it fully utilizes multiple modals of OD flow data as auxiliary information and carefully fuse information at local and global levels for more accurate OD flow predictions.

     \subsubsection{Model Efficiency Analysis}
	In this section, we study the model efficiency of our framework. The training times of all models are shown in Table \ref{tab:computation_time}. All experiments are conducted with the default parameter configurations on a single NVIDIA GeForce P100. We observe that our model had good prediction accuracy and running speed in several of the best-performing baselines. Our framework is superior to most comparison methods and can achieve competitive efficiency compared to DGDR. Modeling spatial dependencies of multiple modal data has higher computational costs though. Considering the prediction accuracy comparison between our framework and DGDR, positive results can be obtained by explicitly utilizing  transportation information of multipel modes. 
	\begin{table}[]
	\centering
 	\caption{Model Efficiency Study With State-of-the-art Deep Learning Methods.}
        \label{tab:computation_time}
	\begin{tabular}{cccc}
		\toprule
		\multirow{2}{*}{Methods} & \multicolumn{3}{c}{Training Time(s)} \\ \cmidrule{2-4} 
		& Taxi       & Bike       & Bus        \\ \midrule
		GCRN                     & 2598.56    & 2249.87    & 2386.51    \\
		GEML                     & 1736.02    & 1437.89    & 1620.95    \\
		DGDR                     & 1229.87    & 1034.73    & 1124.67    \\
		CMOD                     & 1453.63    & 1217.56    & 1329.80    \\ \midrule
		\textsf{FusionTransNet}             & 1289.76    & 1120.53    & 1345.04    \\ \midrule
	\end{tabular}

    \end{table}

	\subsection{Ablation Experiment}
 \label{subsec:ablation}
\begin{figure}[!htbp]
		\centering
  \includegraphics[width=.95\linewidth]{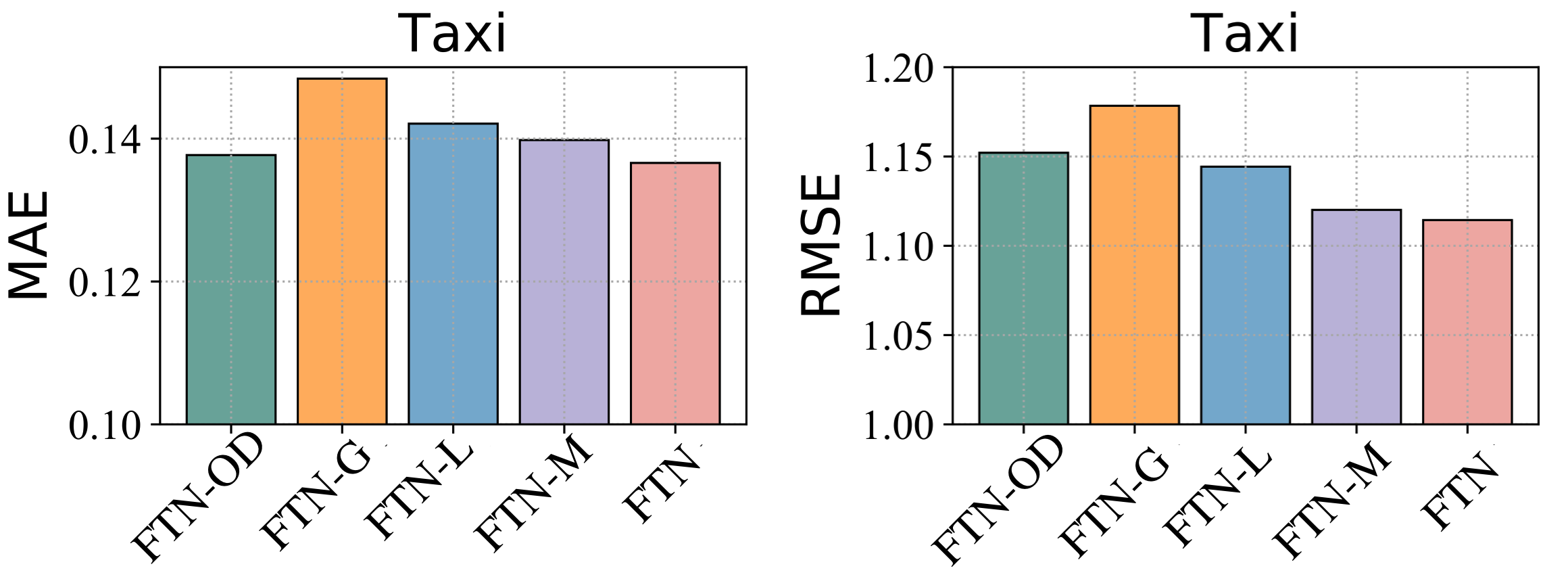}
		
		\caption{Ablation analysis of each component in the proposed framework. 
        We abbreviate \textsf{FusionTransNet} as FTN on the $x$-axes. The first four bars represent ablation versions of our method, with the final bar indicating our complete method. 
  }
        \label{fig:ablation_study}
	   \end{figure}

 To understand the contribution of each component within our \textsf{FusionTransNet} framework, we carried out a series of ablation studies in Figure~\ref{fig:ablation_study}. Using the Shenzhen-taxi dataset as a case study, our aim was to demonstrate how each module affects the overall prediction performance.
 We add variations to \textsf{FusionTransNet} to eliminate each of the important component from the model, including (1) differentiating the roles of origins and destinations to allow for directional flows; (2) the global fusion strategy; (3) the local fusion strategy; and (4) the multiple perspective interaction module.
Below is a breakdown of each variant and its purpose, and its discrepancies in predictive performance with our \textsf{FusionTransNet}.
 	\begin{itemize}
		\item \textbf{FusionTransNet-OD} (same embeddings for both the origins and destinations). 
        This variant employs a traditional GCN in place of our proposed OD-STGCN to gauge the significance of the OD-STGCN over traditional GCN methodologies. The graph structure for GCN is constructed upon the geographic location of stations.
        The substitution of a traditional GCN in place of the OD-STGCN hinders the optimal flow modeling between origins and destinations. 
        By adopting two directed graphs, the OD-STGCN encapsulates the richness of traffic intensities with greater fidelity. This outcome, achieving the second worst in RMSE, demonstrates the salience of OD-STGCN in capturing detailed inter-node traffic flow dynamics.

		\item \textbf{FusionTransNet-G} (the global fusion strategy is removed). To assess the contribution of the global-fusion strategy in modeling inter-modal spatial dependencies, the global-fusion strategy is excluded in this variant, leaving only the local-fusion strategy. 
        This model's performance is the worst for both MAE and RMSE. 
        The worsened performance  in prediction efficacy without the global-fusion strategy accentuated the importance of such a strategy across all components analyzed in our ablation study. The exclusion of global-fusion resulted in a model that was myopic, missing out on capturing broader correlations inherent between multimodal transportation data. 
  
		\item \textbf{FusionTransNet-L} (the local fusion strategy is removed).  To investigate the exclusive impact of the local-fusion strategy on the overall model's performance, local-fusion strategy is eliminated in this setup. 
        The increased errors (the second-worst in MAE), when compared to our full model, indicate the important role of local diffusion. While global-fusion captures overarching correlations across modalities, it potentially overlooks  transitions and conversions at a more granular level. This outcome elucidates the necessity of combining both global and local perspectives for a more holistic model.   
  		\item \textbf{FusionTransNet-M} (the multiple perspective interaction module is removed).  To evaluate the import of the multiple perspective interaction module on our model's predictive performance, we exclude this module (Equation~\eqref{eq:multiple_perspec_1} and Equation~\eqref{eq:multiple_perspec_2}) in this variant, = straightforwardly concatenating output from both fusion strategies. 
        Resorting to mere vector splicing for information fusion fails to encapsulate the dynamism of multiple perspectives. Integrating diverse perspectives is not merely about assimilating information; it is about extracting relevant local and global dependencies that most accurately predict the focal center, which straightforward vector concatenation might overlook.    
	\end{itemize}

    \textbf{FusionTransNet} describes our multimodal spatiotemporal learning framework for OD flow prediction, incorporating all the aforementioned components, to benchmark the above-mentioned variants. 
    In essence, the ablation study affirms the well-calibrated integration of \textsf{FusionTransNet}'s modules, with each playing a pivotal role in sculpting the model's predictive advantage.

    \subsection{Model Interpretations of Inter-modal Learning}
     \label{subsec:interpretations}

    In this section, we analyze the three key components in inter-modal learning, including the local-fusion and global-fusion strategies, as well as the multiple perspective interaction module. 


    \subsubsection{{Effectiveness of Global-fusion Strategy}}
    
\begin{figure*}[htbp]
		\centering
		
		\subfloat{
			\centering
			\includegraphics[width=0.33\linewidth]{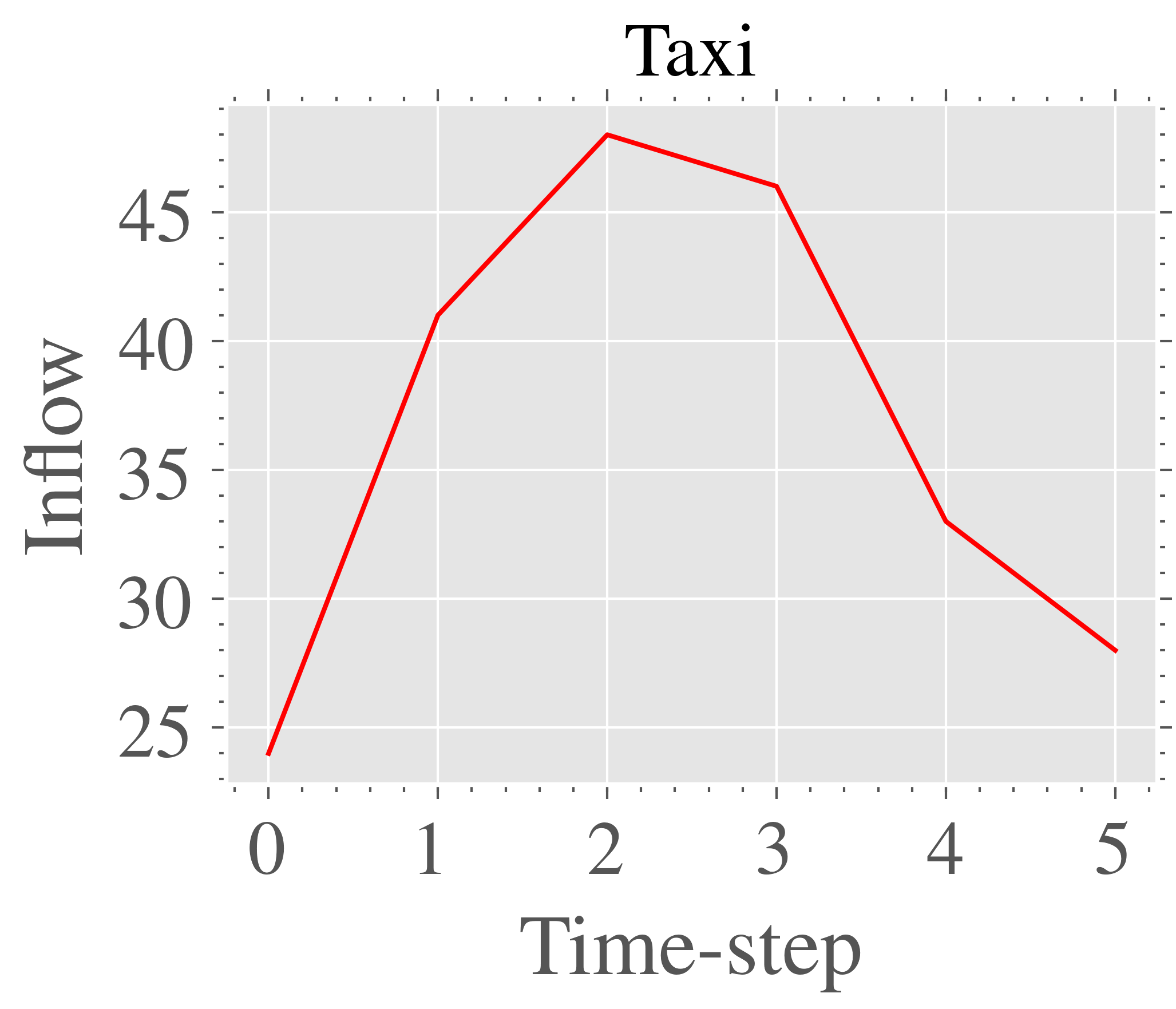}
		}
		\subfloat{
			\centering
			\includegraphics[width=0.33\linewidth]{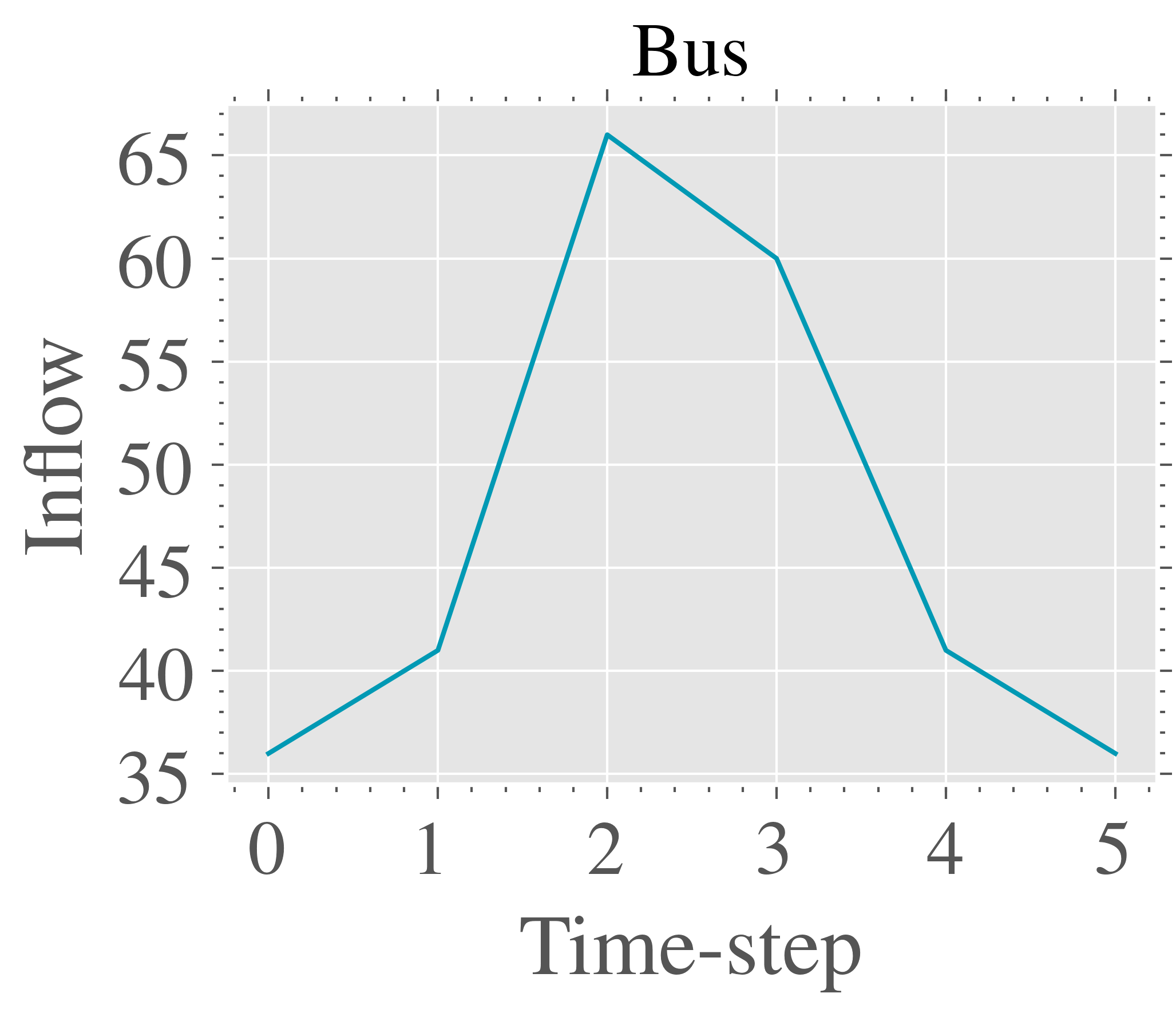}
		}
		\subfloat{
			\centering
			\includegraphics[width=0.33\linewidth]{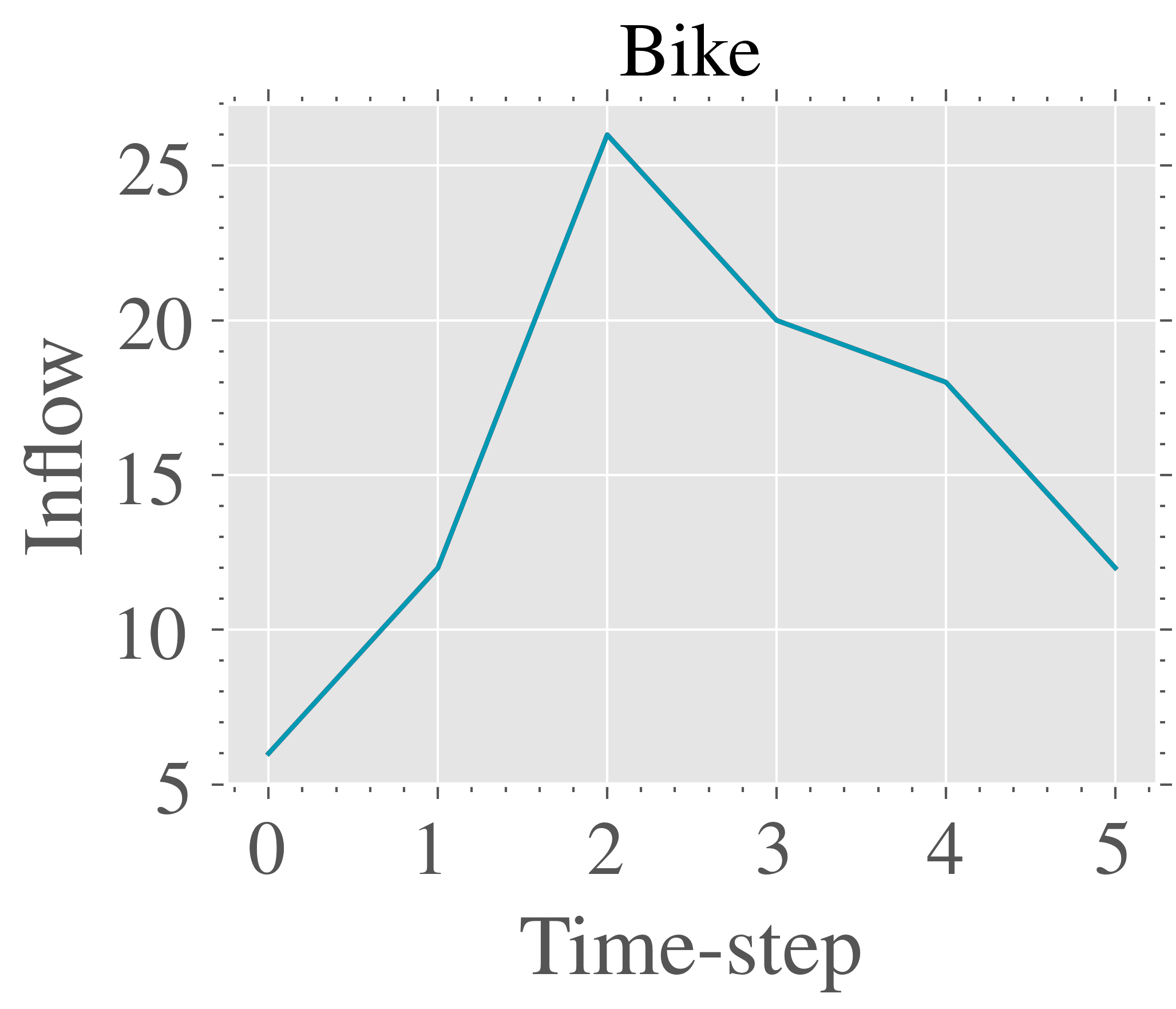}
		} \\ 
  		\subfloat{
			\centering
			\includegraphics[width=0.33\linewidth]{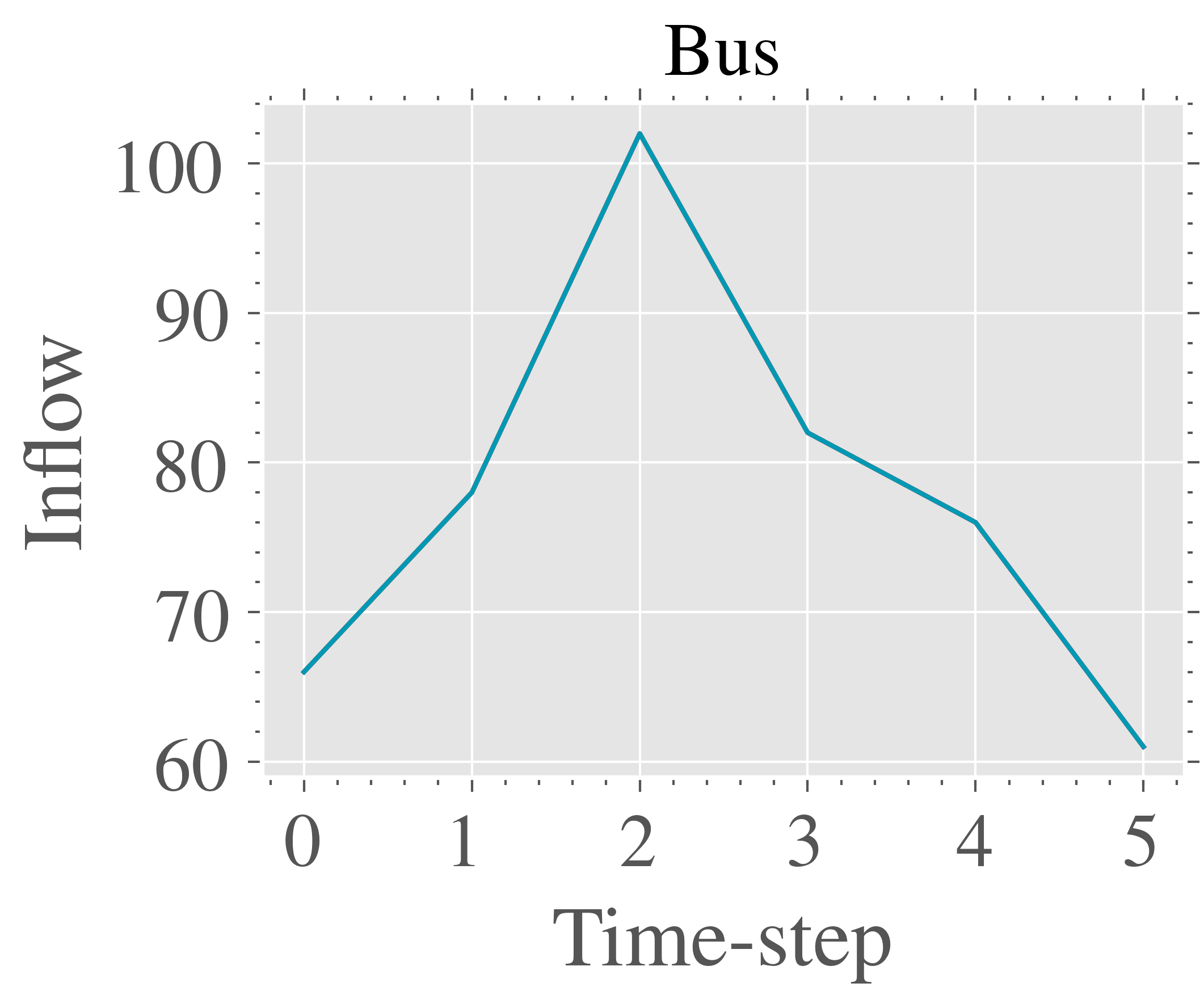}
		}%
  		\subfloat{
			\centering
			\includegraphics[width=0.33\linewidth]{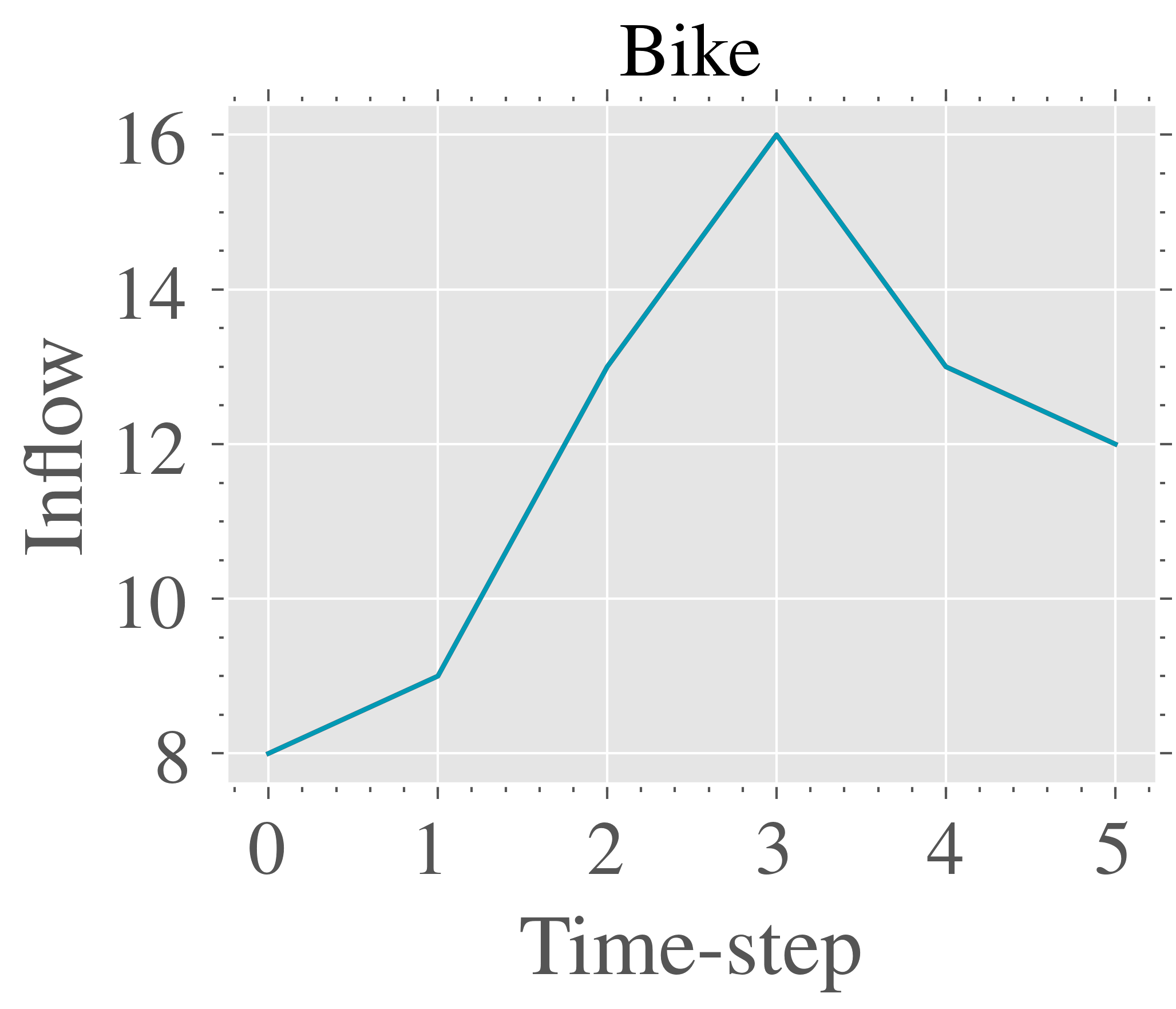}
		}%
		\caption{A case study of the global-fusion strategy (the red line shows the flow of the anchor).}
        \label{fig:interpretation_global}
	\end{figure*}

    We first analyze whether inter-modal nodes with high attention weights are indicative of similar demand patterns. This insight is instrumental for the model to leverage cross-modal data, enabling a more accurate prediction of OD flows that reflects the subtlety of urban transport dynamics.
    In our analysis, we selected the taxi inflow at a particular node as an anchor point (which is the ground truth) and subsequently identified four inter-modal nodes with the highest attention weights, as determined by the global-fusion strategy. Figure 4 presents the actual inflow patterns for both the anchor node (depicted in the first subfigure with a red line) and the selected inter-modal nodes.
    A close inspection of Figure \ref{fig:interpretation_global} reveals a discernible correlation in the inflow patterns among the inter-modal nodes, particularly during peak periods. This similarity in traffic trends, indicated by the synchronized peaks and troughs across the transportation modes, demonstrates the effectiveness of the global-fusion strategy in sensitively identifying nodes with comparable flow demands.
    
    For instance, the peak in taxi inflow at the anchor node during the third time step appears to coincide with inflow peaks in the corresponding bus and bike nodes, albeit at different magnitudes. This pattern suggests a common underlying factor influencing transportation dynamics, such as a city-wide event or a systemic shift in commuter behavior at that time. 
    The model capitalizes on this inter-modal correlation by integrating the shared features from these nodes, which exhibit high similarity in flow patterns. By doing so, it can more accurately infer the traffic state of the anchor node. This ability to assimilate correlated features across modes is particularly useful for improving the precision of OD flow predictions, as it accounts for the complex, interconnected nature of urban transportation systems.

    \subsubsection{{Effectiveness of Local-fusion Strategy}} 
\begin{figure*}[htbp]
		\centering
		
		\subfloat[Transfer-station \#1 \label{fig:interpretation_lobal_a}]{
			\centering
			\includegraphics[width=0.33\linewidth]{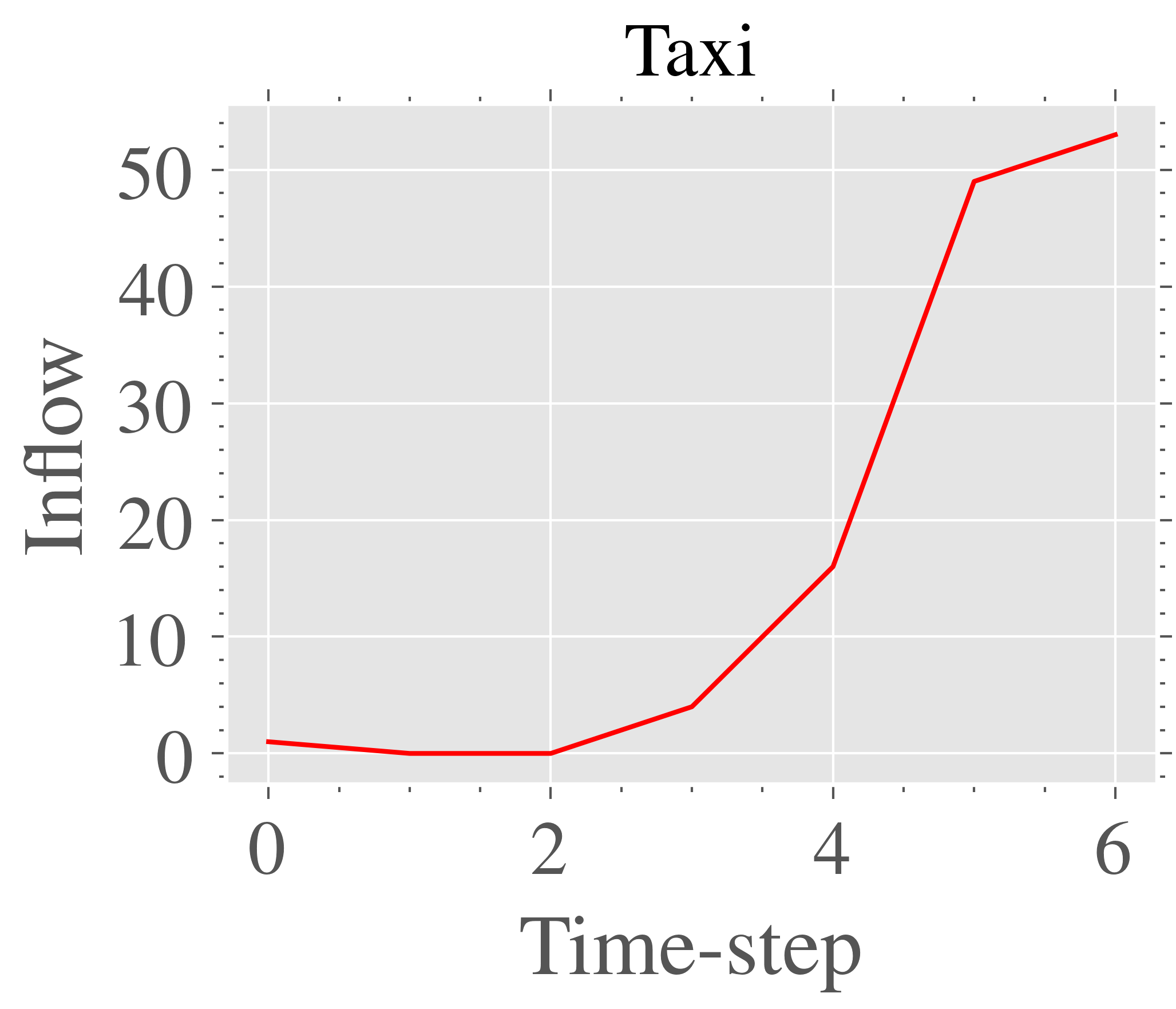}
   			\includegraphics[width=0.33\linewidth]{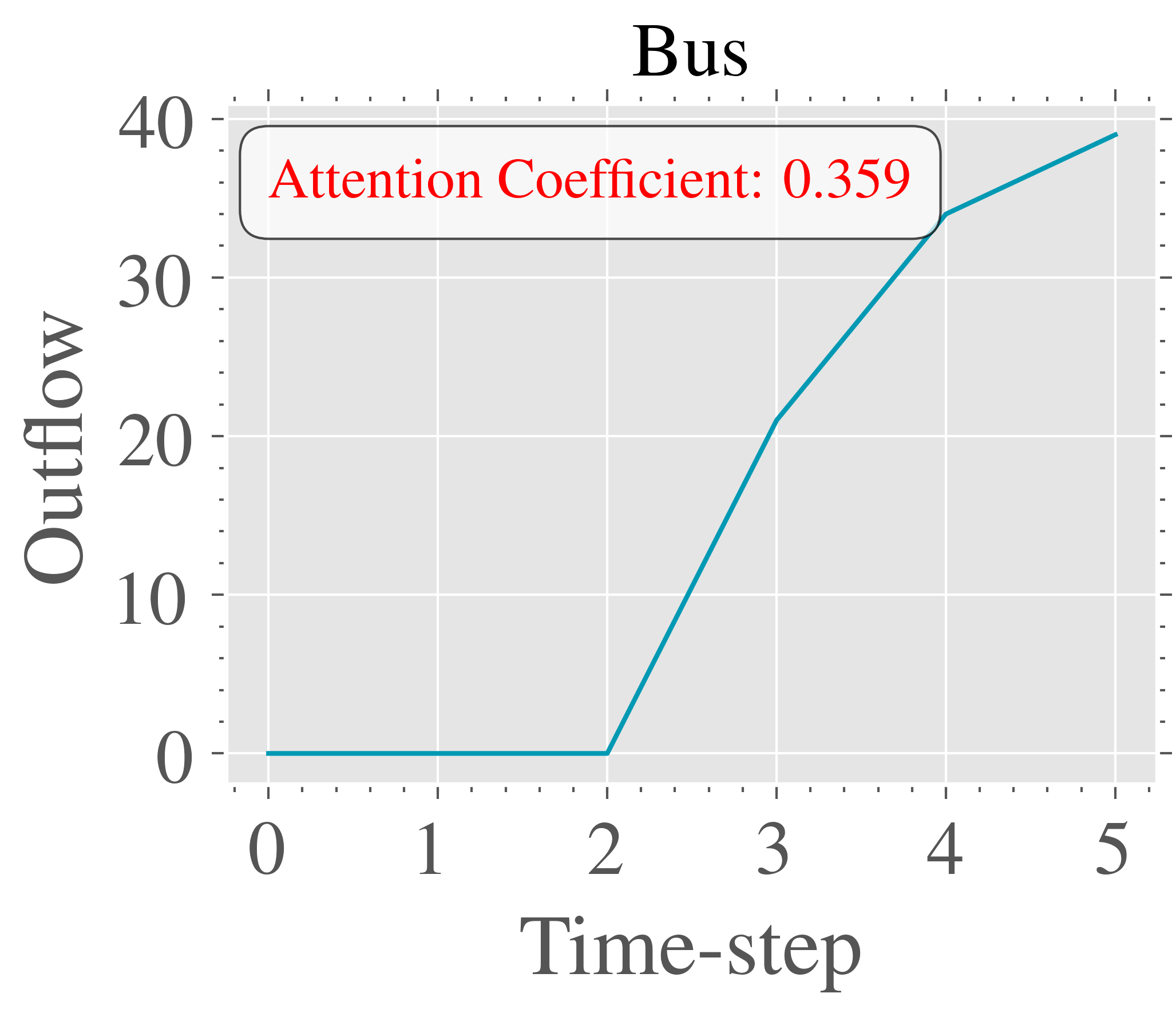}
                \includegraphics[width=0.33\linewidth]{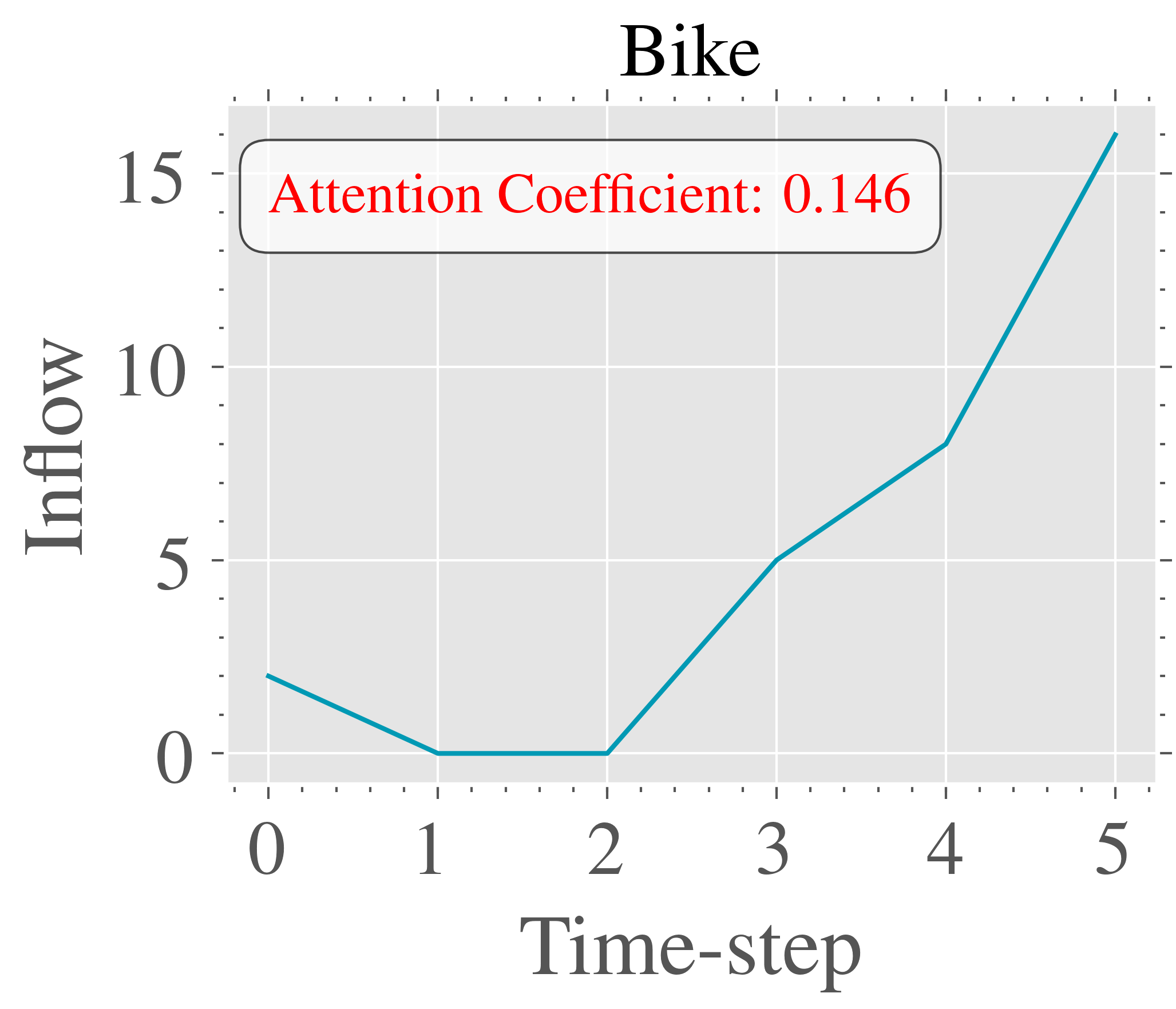}
		}\\
  		\subfloat[Transfer-station \#2 \label{fig:interpretation_lobal_b}]{
			\centering
			\includegraphics[width=0.33\linewidth]{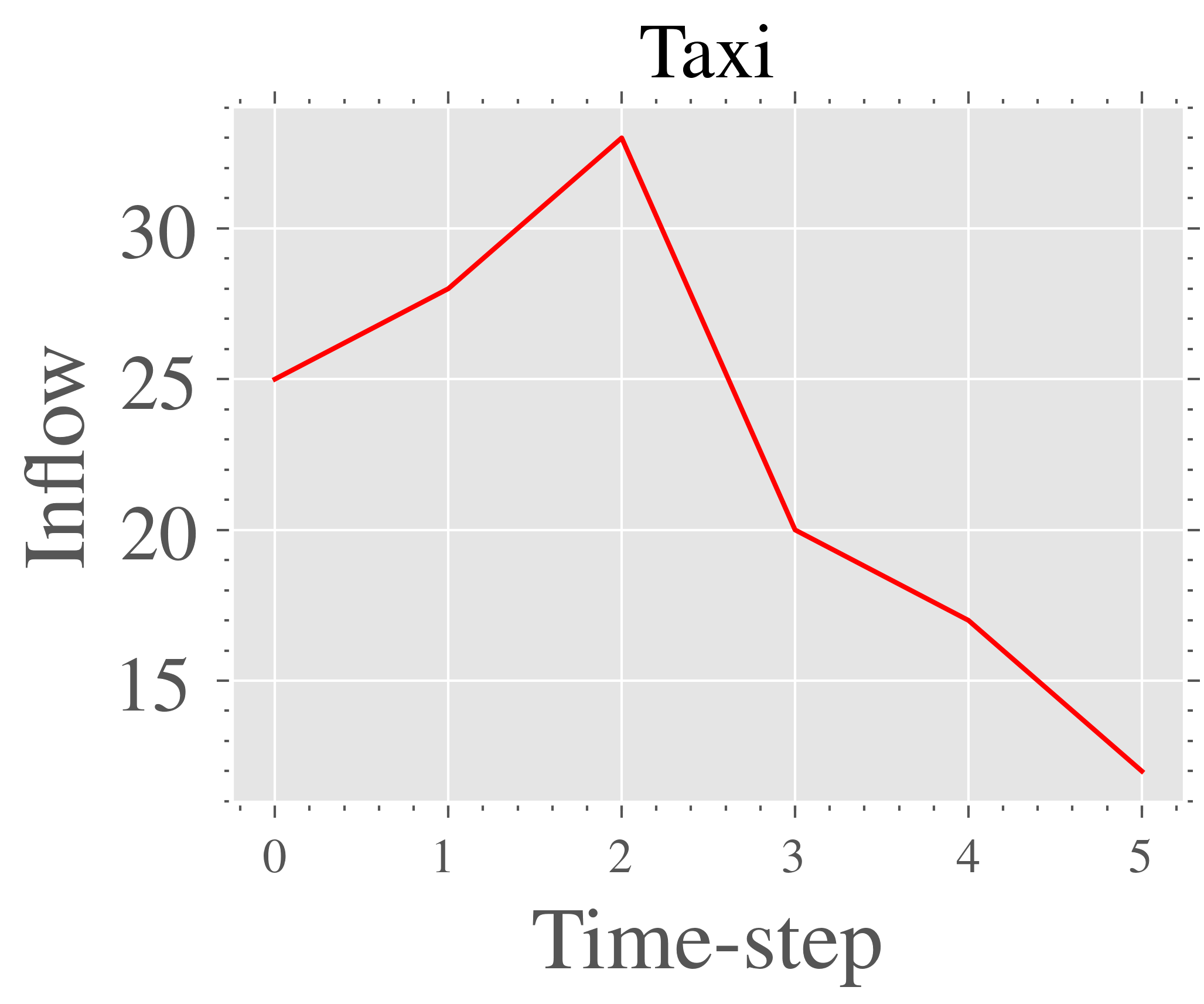}
   			\includegraphics[width=0.33\linewidth]{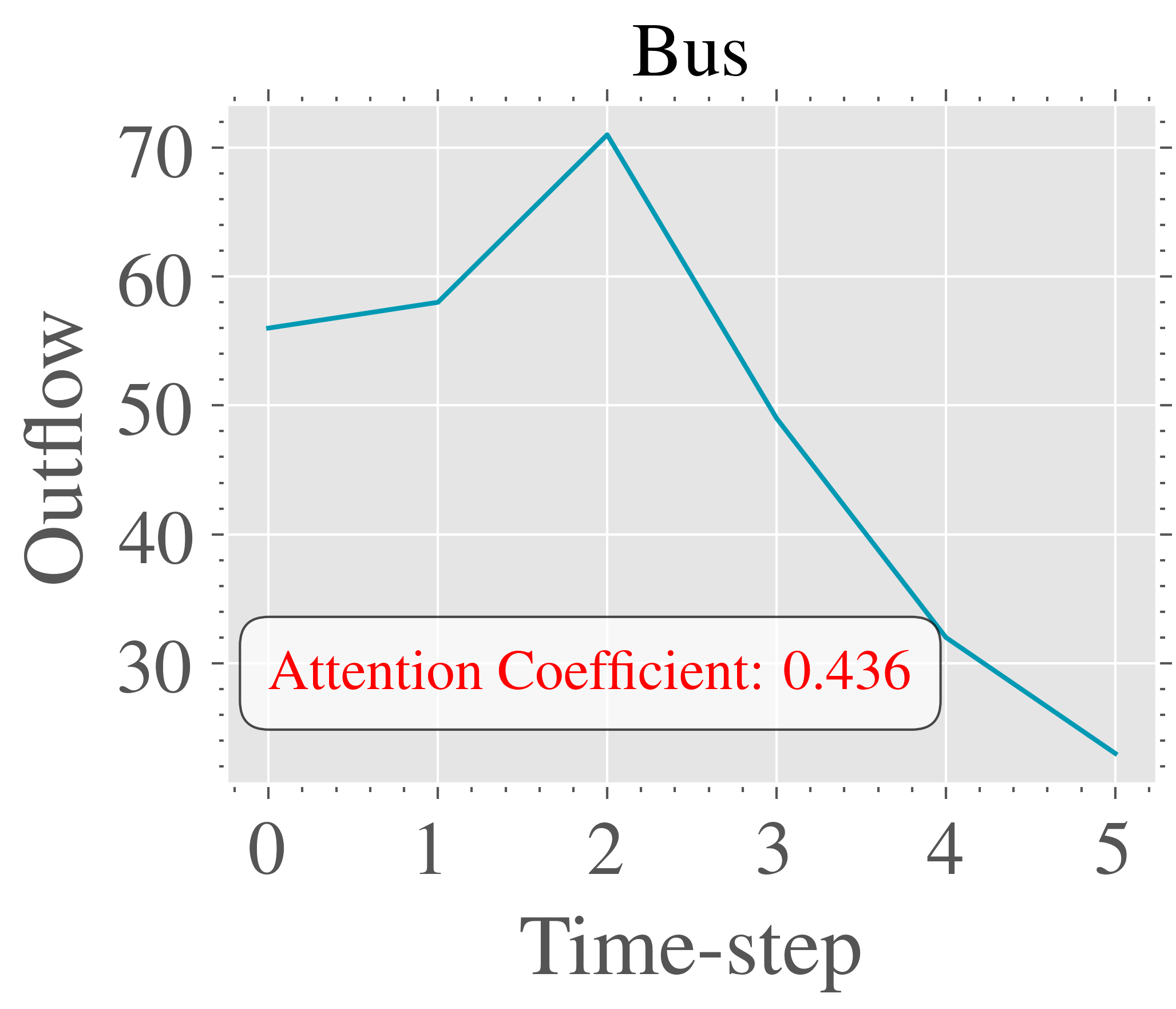}
                \includegraphics[width=0.33\linewidth]{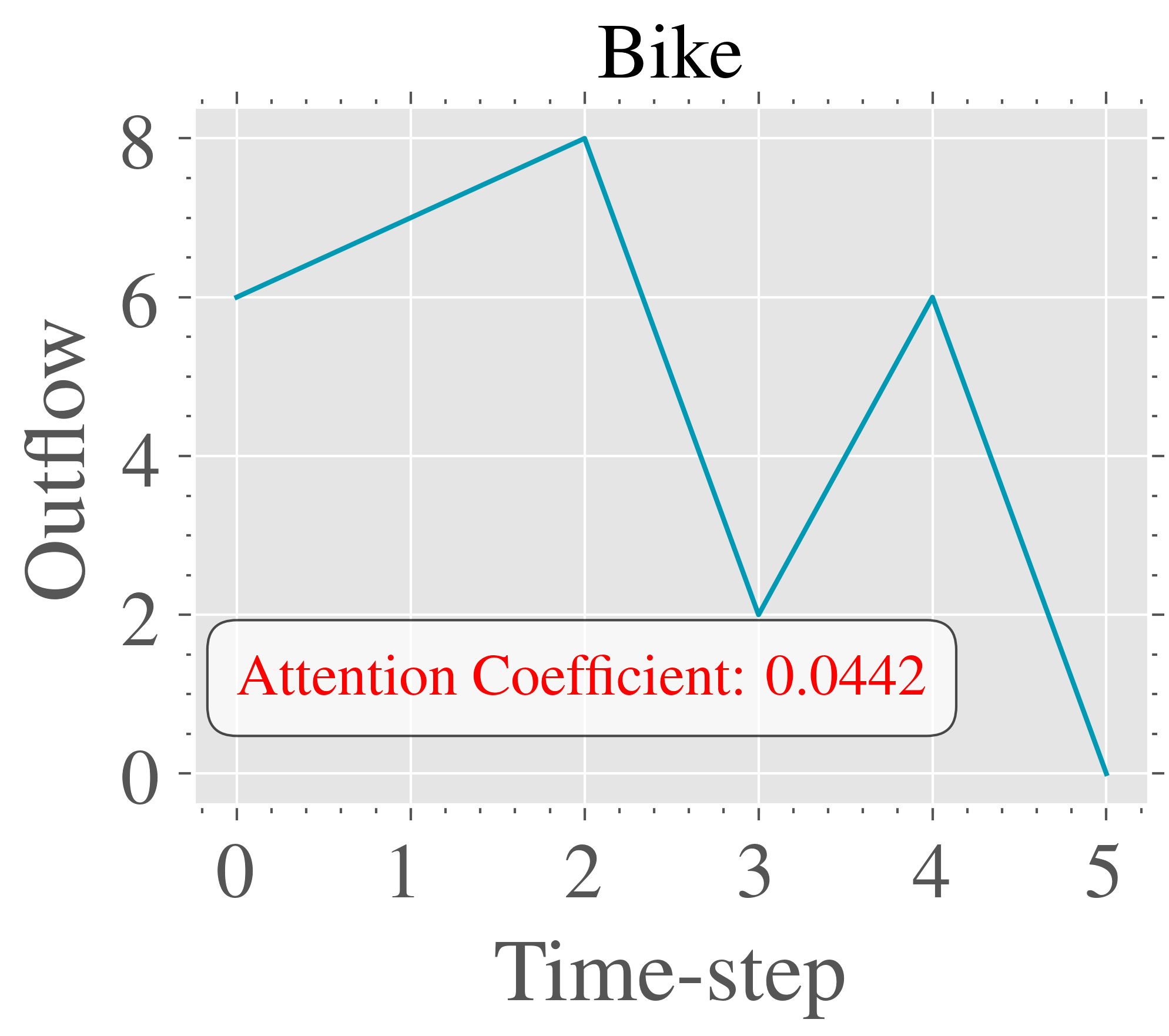}
		}
		\caption{A case study of the local-fusion strategy (the red line shows the flow of the anchor).}
            \label{fig:interpretation_lobal}
	\end{figure*}

    We next analyze the local fusion strategy. 
    The goal is to understand whether the model can construct an accurate representation of dynamic traffic states at each spatial unit. 
    In this analysis, we anchor our focus on the taxi inflow for two distinct nodes to explore the interconnected outflow patterns of other transportation modes facilitated by these transit points. Utilizing the attention coefficients derived from the local-fusion strategy, Figure~\ref{fig:interpretation_lobal} illustrates the outflow distributions for additional modes of transportation, such as buses and bikes, that are associated with the selected taxi nodes.
    
    For example, in Figure~\ref{fig:interpretation_lobal_a}, the taxi node demonstrates a sharp increase in inflow, which is mirrored by a significant outflow in the bus modality, as underscored by the attention coefficient of 0.359. This suggests a high level of interdependence between taxi inflows and bus outflows at this transfer station. On the other hand, the bike modality shows a more gradual increase, reflected by a smaller attention coefficient of 0.146, indicating a less pronounced connection to the taxi inflow at this particular node. Figure~\ref{fig:interpretation_lobal_b} provides a contrasting scenario where a different taxi node exhibits a peak, followed by a sharp decrease in inflow, which is captured with an attention coefficient of 0.436 for the bus outflow. The bike outflow at this node, however, does not correlate strongly with the taxi inflow, as reflected by the even smaller attention coefficient of 0.0442.

    This analysis reveals that the local-fusion strategy adeptly captures the dependencies that exist across modalities, notably between the origins and destinations within the network. The strategy assigns higher attention weights to nodes that display correlated outflow patterns, indicative of a strong inter-modal connection. Conversely, nodes with divergent traffic patterns receive lower attention weights, effectively minimizing the influence of what might be considered noise in the predictive model.

    \subsubsection{Validating Spatial Unit Embedding Against Traffic Sequences} 
    To assess the efficacy of a deep learning method in understanding urban traffic, it is essential that the representation of nodes accurately mirrors actual traffic patterns~\citep{pan2019urban}. To validate the knowledge captured by our model, we identify the k-nearest neighbors for each node within the embedding space. Following this, we evaluate the traffic sequence similarity between the node and its neighbors to determine the representational accuracy of the node embeddings in reflecting real-world traffic behaviors. This step is critical for confirming that the model's learned representations effectively encapsulates traffic similarities.     
    In our evaluation process, two metrics were employed: Pearson correlations (POC) and first-order temporal correlations (FOC) ~\citep{chouakria2007adaptive}. These metrics are useful in determining the affinity between two temporal sequences:

 \begin{equation}
		\begin{aligned}
			&\operatorname{POC}(\mathbf{m}, \mathbf{n})=\frac{\sum_{i}\left(m_{i}-m_{i-1}\right)\left(n_{i}-n_{i-1}\right)}{\sqrt{\sum_{i}\left(m_{i}-m_{i-1}\right)^{2}} \sqrt{\sum_{i}\left(n_{i}-n_{i-1}\right)^{2}}}, \\
   			&\operatorname{FOC}(\mathbf{m}, \mathbf{n})=\frac{\sum_{i}\left(m_{i}-\bar{m}\right)\left(n_{i}-\bar{n}\right)}{\sqrt{\sum_{i}\left(m_{i}-\bar{m}\right)^{2}} \sqrt{\sum_{i}\left(n_{i}-\bar{n}\right)^{2}}}, 
		\end{aligned}
	\end{equation}
    where $\mathbf{m}$ and $\mathbf{n}$ indices two sequences of nodes, $\bar{m}$ and $\bar{n}$ are their mean values.

\begin{figure}[htbp]
	\centering
 \includegraphics[width=.95\linewidth]{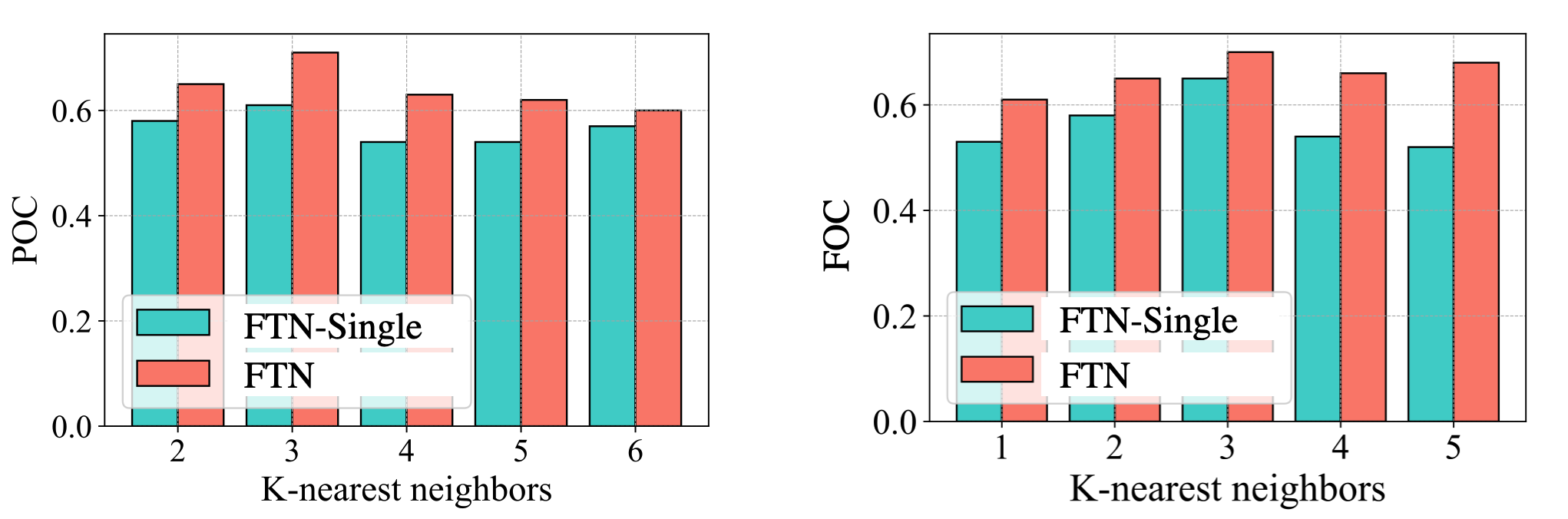}

	\caption{Evaluation on flow similarity between each node and its k-nearest neighbors in the embedding space. We abbreviate \textsf{FusionTransNet} as FTN in the legend. }
	\label{fig:evaluate_MPI}
\end{figure}

    Figure~\ref{fig:evaluate_MPI} compares the performance of our proposed \textsf{FusionTransNet} model against \textsf{FusionTransNet}-Single, which utilizes single-modal information. The results are quantified using the FOC and POC metric across varying numbers of k-nearest neighbors.
    The results show that \textsf{FusionTransNet} outperforms \textsf{FusionTransNet}-Single in both metrics across all considered k-nearest neighbor counts. The higher FOC and POC values indicate that \textsf{FusionTransNet}, with its multimodal approach, places nodes with similar traffic demand patterns closer together in the embedding space. This implies that \textsf{FusionTransNet} embeddings better represent the actual similarities in traffic flows between nodes compared to \textsf{FusionTransNet}-Single.
    The consistency of higher FOC and POC values for \textsf{FusionTransNet}, regardless of the number of k-nearest neighbors examined, suggests a robustness in the model's ability to group similar nodes.

     The comprehensive embedding space learned from \textsf{FusionTransNet} is advantageous for predicting OD flow since it encapsulates a more detailed understanding of the network’s traffic dynamics. This multimodal integration allows for a richer and more accurat representation of traffic flows, which is key for predicting how traffic will move through the urban network.

	\section{Conclusion}

In this study, we introduced \textsf{FusionTransNet}, a comprehensive framework designed to enhance the prediction of OD flow within complex urban transportation networks. By leveraging multimodal data, \textsf{FusionTransNet} effectively captures the dynamic interactions between different transportation modes, offering a detailed representation of urban mobility patterns. Our approach integrates both global and local perspectives on traffic flow, utilizing a novel combination of global fusion, local fusion, and multiple perspective interaction modules. This enables the model to discern nuanced traffic behaviors across various modalities, from the overarching trends affecting the entire network down to the intricate dynamics at individual multi-modal stations. Through empirical evaluations in metropolitan settings, including Shenzhen and New York, \textsf{FusionTransNet} demonstrated superior predictive accuracy compared to existing models, underscoring the significant benefits of incorporating multimodal information into urban transportation modeling.

The ablation studies demonstrated the significance of each component within \textsf{FusionTransNet}, showing that removing any key feature—be it the differentiation between origins and destinations, the global fusion strategy, the local fusion strategy, or the multiple perspective interaction module—detrimentally affects the model's performance. Specifically, the experiments highlighted the critical roles of the global and local fusion strategies in capturing broad traffic trends and detailed modal interactions, respectively. The diminished predictive performance observed in the absence of these components underscores their collective importance in modeling the complexity of urban transportation systems.

Simultaneously, our interpretations of how \textsf{FusionTransNet} utilizes these components to model traffic dynamics further validate their efficacy. The global fusion strategy's ability to discern inter-modal correlations, especially during peak traffic periods, enhances the model's grasp of city-wide traffic flows. In parallel, the local fusion strategy's focus on the interactions between different transportation modes at specific nodes,  such as individual stations or neighborhoods, enriches the model with granular insights into traffic behavior, reflecting its necessity for accurate local traffic state predictions. 
Furthermore, the evaluation of spatial unit embedding against traffic sequences, using metrics such as FOC and POC, reinforces the model's capacity to accurately mirror real-world traffic patterns. 
This comprehensive modeling of traffic flows, enabled by \textsf{FusionTransNet}, not only establishes a new standard for traffic flow modeling but also carries profound implications for urban planning, traffic management, and beyond.

\paragraph{Theoretical Implications}
Our approach illustrates the efficacy of combining localized and systemic analyses for a richer understanding and prediction of complex, multi-modal systems. By incorporating both local and global fusion strategies, complemented by a multiple perspective integration module, we present a structured framework for analyzing multi-modal systems. This methodology allows for an in-depth exploration of interactions at various scales, from the micro-dynamics at individual stations or neighborhoods to macro-scale trends affecting the entire network. Such a comprehensive model addresses previous shortcomings by capturing the influence of both localized conditions and broader systemic patterns on urban mobility. Furthermore, this conceptual framework for data fusion, exemplifying the synergy between local and global insights, has broader applicability in dissecting other complex systems with multiple interacting components, like supply chain logistics or energy distribution networks. This versatility underscores the model's potential to inform theory and practice in various domains.

\paragraph{Practical  Implications}
On the practical front, our findings highlight the utility of leveraging existing urban infrastructure data—Global Positioning System (GPS), smart card transactions, mobile payments—to fuel our model. This accessibility allows municipal authorities to adopt our \textsf{FusionTransNet} framework without the need for significant new technological investments. More than just forecasting, our model's architecture is poised to enhance traffic anomaly detection, event response management, and service optimization. Its adaptability to both present conditions and predictive analytics for future scenarios offers cities an opportunity to advance their transportation systems towards greater efficiency and responsiveness.

\paragraph{Future Work}
 Our work opens up several directions for further exploration. 
 First, integrating \textsf{FusionTransNet} with transportation policymaking could significantly improve urban mobility planning. The model's predictive accuracy offers a tool for designing more efficient and sustainable transportation strategies. By predicting traffic flow patterns accurately, policymakers can make informed decisions on infrastructure improvements and public transit adjustments, potentially reducing traffic congestion and enhancing commuter experiences.
 
Second, another promising direction involves applying \textsf{FusionTransNet} to other complex socio-technical systems characterized by spatiotemporal data. Specifically, the model could be adapted to study the spread of epidemics across different modes of transportation. This application is particularly relevant given the interconnected nature of modern transportation networks and their role in disease transmission. By modeling how an epidemic might spread through these networks, researchers could provide valuable insights into preventive measures and containment strategies. 
In addition, applicability to other complex systems that feature spatiotemporal data is another area for expansion. This includes systems like supply chain logistics, where predicting demand and supply dynamics could optimize distribution routes; energy distribution networks, where forecasting energy demand could lead to more efficient energy use; and telecommunication networks, where understanding data flow patterns could improve network service quality. 
Moreover, this approach reaffirms the importance of considering the socio-technical aspects of systems, where human, systems, and technology interact in complex ways. For IS researchers, this presents an opportunity to explore and contribute to the understanding and management of such systems, extending beyond traditional domains to address critical public health issues. 

 
\ACKNOWLEDGMENT{}





\bibliographystyle{abbrvnat}
\bibliography{aaai222}

\end{document}